%% file: main.tex
\documentclass[10pt,twocolumn,letterpaper]{article}

\usepackage{iccv}              

\input{preamble}

\def\paperID{5354}
\def\confName{ICCV}
\def\confYear{2025}

\title{Deep Space Weather Model: \\Long-Range Solar Flare Prediction from Multi-Wavelength Images}

\author{
Shunya Nagashima \qquad Komei Sugiura\\
Keio University, Japan\\
{\tt\small \{ng\_sh, komei.sugiura\}@keio.jp}
}

\begin{document}
\maketitle
\input{pages/abstract}
\input{pages/section_1}

\input{pages/section_2}
\input{pages/section_3}

\input{pages/section_4}

\input{pages/section_5}
\input{pages/section_6}

\input{pages/section_7}
{
    \small
    \bibliographystyle{ieeenat_fullname}
    \bibliography{reference}
}

\input{pages/suppl}
\end{document}

%% file: preamble.tex
\usepackage{graphicx}
\usepackage{amsmath}
\usepackage{amssymb}
\usepackage{booktabs}
\usepackage{tabularx}
\usepackage{bm}
\usepackage{float}
\usepackage{wrapfig}
\usepackage{calc}
\usepackage{colortbl}
\usepackage{multirow}
\usepackage{tikz}
\usepackage{mdframed}
\usetikzlibrary{positioning}
\usepackage{rotating}
\usepackage{hhline} 
\usepackage{makecell}

\definecolor{iccvblue}{rgb}{0.21,0.49,0.74}
\definecolor{myred}{HTML}{B85450}
\definecolor{myblue}{HTML}{6C8EBF}
\definecolor{polaris}{RGB}{121,150,196}
\definecolor{composite}{RGB}{230,163,126}
\definecolor{flickr8k-cf}{RGB}{129,190,142}
\definecolor{flickr8k-ex}{RGB}{211,123,126}
\definecolor{TitleColor}{gray}{0.95}
\definecolor{LightPink}{HTML}{FAE6E7}

\usepackage[accsupp]{axessibility}

\usepackage[pagebackref,breaklinks,colorlinks,allcolors=iccvblue]{hyperref}

%% file: pages/abstract.tex
\begin{abstract}

Accurate, reliable solar flare prediction is crucial for mitigating potential disruptions to critical infrastructure, while predicting solar flares remains a significant challenge. Existing methods based on heuristic physical features often lack representation learning from solar images. On the other hand, end-to-end learning approaches struggle to model long-range temporal dependencies in solar images.
In this study, we propose Deep Space Weather Model (Deep SWM), which is based on multiple deep state space models for handling both ten-channel solar images and long-range spatio-temporal dependencies. 
Deep SWM also features a sparse masked autoencoder, a novel pretraining strategy that employs a two-phase masking approach to preserve crucial regions such as sunspots while compressing spatial information.
Furthermore, we built FlareBench, a new public benchmark for solar flare prediction covering a full 11-year solar activity cycle, to validate our method.
Our method outperformed baseline methods and even human expert performance on standard metrics in terms of performance and reliability. The project page can be found at \href{https://keio-smilab25.github.io/DeepSWM}{\textcolor{magenta}{https://keio-smilab25.github.io/DeepSWM}}.


\end{abstract}

%% file: pages/section_1.tex
\vspace{-5mm}
\section{Introduction}
\vspace{-2mm}

\begin{figure}[t]
    \centering
    \includegraphics[width=\linewidth]{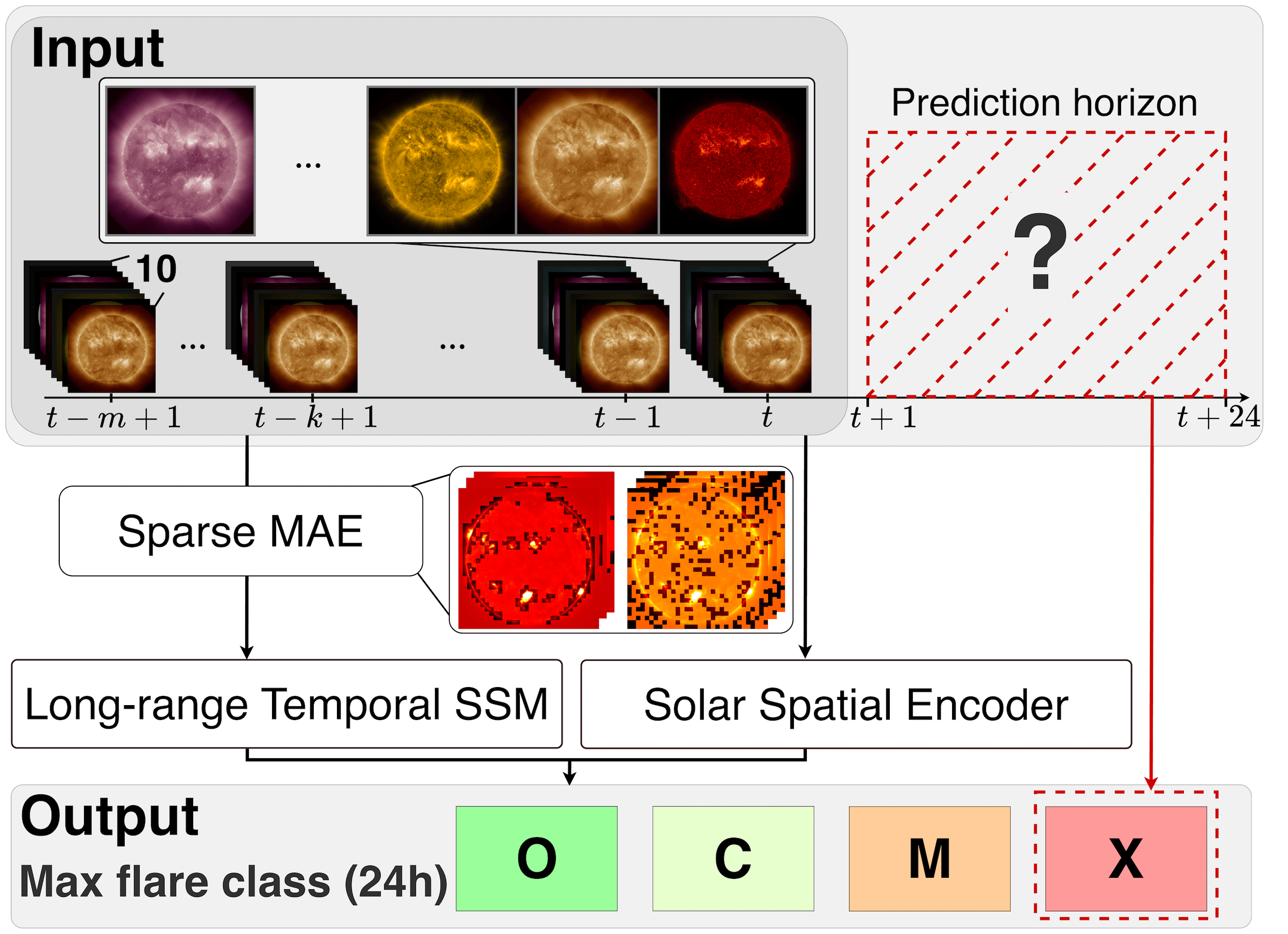}
    \vspace{-5mm}
    \caption{Overview of the proposed method, Deep SWM. The input is a sequence of multi-channel solar images (top). These images are processed by the Solar Spatial Encoder. Concurrently, the images are used for pretraining with a Sparse Masked Autoencoder, and the resulting representations are passed to the Long-range Temporal State Space Model. The outputs of these two modules are combined to output a predicted probability corresponding to one of the four solar flare classes (X, M, C, and O) (bottom).}
    \label{fig:eye-catch}
    \vspace{-3mm}
\end{figure}

Solar flares are intense bursts of electromagnetic radiation that can significantly impact critical infrastructure including GPS systems, communication networks, spacecraft, and power grids \cite{Bhattacharjee_ApJ_2020, Deshmukh_aaai_2021}.
The potential economic damage from a major solar flare is estimated to be 0.6 to 2.6 trillion US dollars \cite{Maynard_techreport_2013}. Accurate and reliable solar flare prediction is therefore crucial for mitigating these potential risks by, for example, rerouting flights, preparing satellites for safe-mode operations, and bolstering power grid resilience.

In this study, we focus on predicting the class of the largest solar flare occurring within 24 hours, formulated as a multi-class classification problem.
Despite decades of research, accurate and reliable solar flare prediction remains a significant challenge,
even for skilled human experts \cite{Kubo_humanperf_2017, Murray_humanperf_2017}.
Human expert performance is quantified by a Gandin-Murphy-Gerrity score (GMGS) \cite{Gandin_gmgs_1992} and a Brier skill score (BSS$_{\scalebox{0.8}{$\geq \mathrm{M}$}}$) \cite{Nishizuka_bss_2020} of 0.48 and 0.16 \cite{Kubo_humanperf_2017}, whereas a perfect forecast would achieve a 1.0 on each metric.

Existing approaches often face challenges in representation learning and temporal modeling. Typical approaches based on heuristic physical features, such as Spaceweather HMI Active Region Patch (SHARP) parameters \cite{Bobra_ApJ_2015, Leka_ApJ_2019, Herdiwijaya_JPCS_2024}, often fail to capture the finer details preceding solar flare eruptions \cite{Barnes_ApJ_2016,Toriumi_livingreviews_2019, Li_AAS_2024} because of the lack of direct representation learning from the images. By contrast, end-to-end approaches \cite{Pandey_BigData_2021, Abed_ASR_2021} (although they learn representations) sometimes fail to adequately model the long-range temporal dependencies critical for accurate prediction.

To address these limitations, we propose Deep Space Weather Model (Deep SWM), an end-to-end solar flare prediction model that (1) learns representations from a time series of solar images with sparse important regions and (2) extends deep state space models (deep SSMs) to capture long-range spatio-temporal dependencies. By learning directly from time series of solar images, our model avoids the limitations of the physical features but captures representations of pre-flare states. Furthermore, by extending deep SSMs, our model accurately captures the long-range spatio-temporal dependencies that are crucial for accurate flare prediction. These dependencies represent complex solar interactions and evolving patterns over time. They are precursors to flares and are often missed by traditional methods focusing on localized features. 

Fig.~\ref{fig:eye-catch} illustrates an overview of Deep SWM. It takes as input a sequence of multi-channel solar images, such as those captured by the SDO/HMI \cite{Scherrer_SoPh_2012} and SDO/AIA \cite{Lemen_SoPh_2012} instruments. It outputs a predicted probability for one of the four flare classes (X, M, C, and O), indicating the largest flare class expected within the next 24 hours. Note that these classes are ordered by peak X-ray flux, with X representing the strongest events. Detailed information on the flare classes can be found in Appendix \ref{supp:flare_class}.


The novelties of this study are outlined as follows:
\begin{itemize}
    \setlength{\parskip}{0.2mm} 
    \setlength{\itemsep}{0.2mm} 
    \item We have constructed a new public benchmark for solar flare prediction, FlareBench, covering a complete 11-year solar activity cycle to mitigate the risk of biased evaluations inherent in shorter-duration datasets.
    \item We propose the Solar Spatial Encoder, which selectively weights channels and captures long-range spatio-temporal dependencies over time series of multi-channel solar images.
    \item We introduce the Long-range Temporal State Space Model, extending deep SSMs to accurately capture long-range temporal dependencies and fine-grained temporal features exceeding the solar rotation period.
    \item We introduce Sparse Masked Autoencoder, a pretraining method tailored for sparse images, improving the representation of crucial, yet sparse, informative regions, such as sunspots. 
    \item We present the first successful attempt to demonstrate that an end-to-end solar flare prediction model can achieve superhuman performance in terms of both GMGS (weighted accuracy) and BSS (reliability).
\end{itemize}

%% file: pages/section_2.tex
\vspace{-1mm}
\section{Related Work}
\vspace{-1mm}
There exist several comprehensive surveys on recent advances in solar flare prediction \cite{Georgoulis_JSWSC_2021}.
In addition, masked autoencoders (MAEs), explored in this study, have been used for pretraining visual models, with successful applications to image representation learning summarized in \cite{Zhang_IJCAI_2023}. Furthermore, deep state space models (deep SSMs), also a key component of our approach, have emerged as a powerful method for sequence modeling, with recent advances comprehensively summarized in \cite{Patro_arxiv_2024, Wang_arxiv_2024}.



\vspace{-3mm}
\paragraph{AI for space.}

Observations from space, coupled with advancements in AI, have significantly enhanced our understanding of astronomical phenomena and have been instrumental in advancing solar physics \cite{Chin_CVPR_Workshops_2019, Felt_WACV_2024, Hu_CVPR_2021, Mesa_CVPRW_2021}. Within this rapidly evolving field, solar flare prediction plays a crucial role in space weather forecasting.
Many methods have been proposed for solar flare prediction \cite{Deshmukh_aaai_2021, Nishizuka_DeFN_2018, Park_ApJ_2018, Liu_ApJ_2019, Kaneda_FlareTransformer_2022, Iida_ACCV_2022, Li_AAS_2024}. 
 Early studies often employed MLPs operating on extracted physical features \cite{Nishizuka_DeFN_2018, Florios_SoPh_2018}, such as DeFN \cite{Nishizuka_DeFN_2018}. 
Recently, transformer-based models have been explored for their ability to capture long-range dependencies \cite{Kaneda_FlareTransformer_2022, Abduallah_SolarFlareNet_2023, Grim_SoPh_2024, Li_AAS_2024}. The Flare Transformer \cite{Kaneda_FlareTransformer_2022} models temporal relationships between magnetograms and physical features.


\vspace{-2mm}
\paragraph{Masked autoencoders.}
\vspace{-1mm}
Masked image modeling has emerged as a powerful paradigm for self-supervised learning in computer vision, drawing inspiration from the success of masked language modeling \cite{Sinha_EMNLP_2021, Barba_ACL_2023, Min_ACL_2023}. The seminal work on MAE \cite{He_MAE_2022} demonstrated that a simple yet effective strategy of masking random patches of an input image and reconstructing the missing pixels could yield scalable and robust visual representations. The core of MAE's design lies in its asymmetric encoder-decoder architecture; the encoder processes only visible patches, whereas the decoder reconstructs the entire image from the latent representation and mask tokens. Critically, \cite{He_MAE_2022} found that a high masking ratio (e.g., 75\%) usually creates a non-trivial self-supervisory task, forcing the model to learn holistic image understanding.

\begin{figure*}[t]
    \centering
    \includegraphics[width=0.88\linewidth]{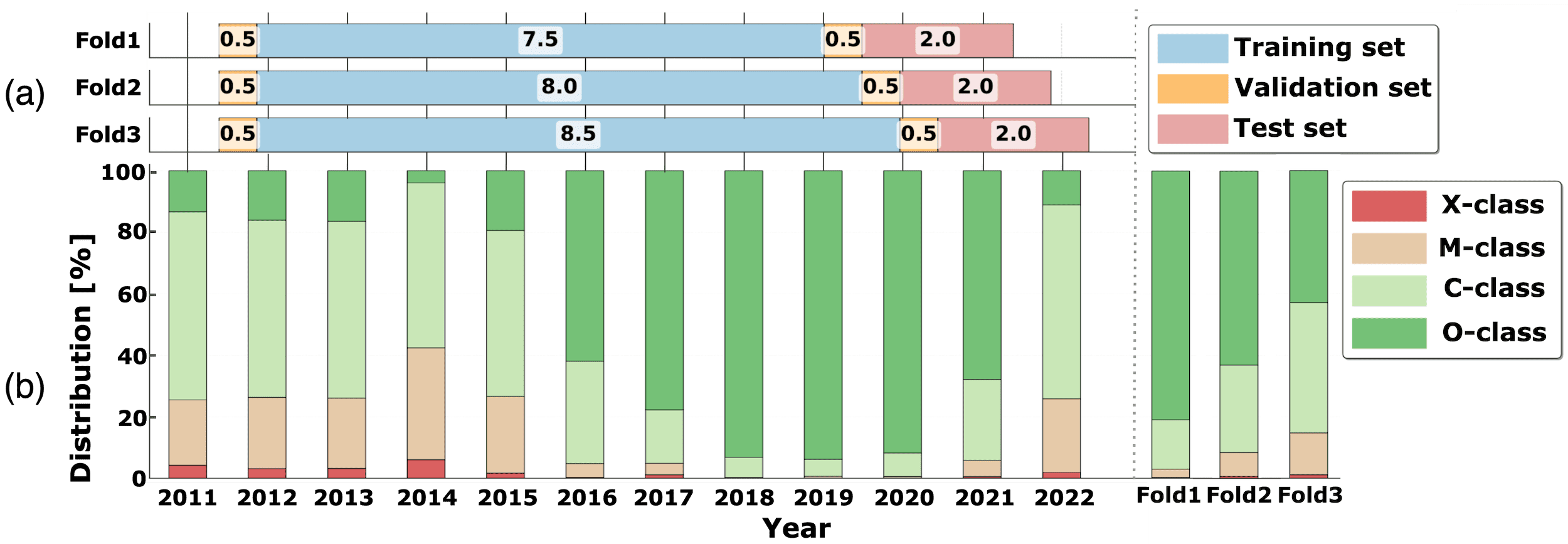}
    \vspace{-3mm}
    \caption{(a) Dataset division for time-series cross-validation. The dataset is divided into three folds (Fold 1, Fold 2, and Fold 3), each comprising a training set, a validation set, and a test set. The numbers within each bar denote the duration (in years) of the corresponding periods. (b) Annual distribution of solar flare classes.}
    \label{fig:dist_aplit}
    \vspace{-2mm}
\end{figure*}

\vspace{-2mm}
\paragraph{Deep SSMs.}
\vspace{-1mm}
While the Transformer architecture \cite{Vaswani_NeurIPS_2017} has become standard in various fields \cite{Brown_NeurIPS_2020, Baevski_NeurIPS_2020, Dosovitskiy_ICLR_2021}, its core attention mechanism suffers from quadratic computational complexity concerning sequence length, posing scalability challenges \cite{Patro_arxiv_2024}. Consequently, considerable research has focused on architectures that maintain the representational power of Transformers while reducing computational cost, particularly for long sequences \cite{Ren_NeurIPS_2021, Han_ICCV_2023}. Deep SSMs \cite{Gu_ICLR_2022, Smith_S5_2023, Gu_Mamba_2024} have attracted significant research interest, given their ability to model long-range dependencies.
Building upon foundational work in deep SSMs, S5 \cite{Smith_S5_2023} uses a multi-input, multi-output SSM and an efficient parallel scan for computation, improving the computational efficiency of deep SSMs.
Mamba \cite{Gu_Mamba_2024} introduces a selection mechanism that enables a time-varying deep SSM, leading to reported improvements over Transformers in natural language processing and spurring significant interest. 

Appendix \ref{supp:related_work} explains other related studies.

%% file: pages/section_3.tex
\vspace{-1mm}
\section{FlareBench}
\label{seq:flarebench}
\vspace{-1mm}
\paragraph{Why do we need a new benchmark?}
Most conventional datasets for solar flare prediction \cite{Nishizuka_ApJ_2017, Angryk_ScientificData_2020} do not cover diverse solar activity states.
Consequently, models trained on such datasets can exhibit biases towards specific periods of solar activity. Furthermore, many datasets contain only small, low-resolution sunspot patches. To address these limitations, we propose FlareBench, a novel benchmark for solar flare prediction. FlareBench covers the entire 11-year solar activity cycle for evaluating models under diverse solar activity states.
FlareBench focuses on predicting the maximum class of solar flare within the next 24 hours, a standard approach in solar flare prediction \cite{Nishizuka_ApJ_2017, Leka_ApJ_2019, Zheng_ApJ_2019}. Furthermore, time-series cross-validation is used for model evaluation to ensure that results are not biased towards specific periods. 

FlareBench presents the following unique challenges:
\begin{itemize}
    \item It requires modeling long-term, diverse solar states spanning the entire 11-year solar cycle.
    \item Computationally efficient architectures are necessary to model multi-wavelength images that capture multi-layered physical phenomena across various atmospheric layers.
\end{itemize}

\vspace{-4mm}
\paragraph{Dataset composition.}
Building a benchmark that covers the entire 11-year solar activity cycle has only recently become possible for the following reasons.
The FlareBench dataset comprises continuous multi-wavelength solar observation data, acquired by the Helioseismic and Magnetic Imager (HMI) \cite{Scherrer_SoPh_2012} and Atmospheric Imaging Assembly (AIA)\cite{Lemen_SoPh_2012} instruments onboard the SDO\cite{Pesnell_SDO_2012}. SDO/HMI began observations on May 1, 2010, and SDO/AIA on April 28, 2010, respectively. 
The availability of this full-cycle dataset, combining both HMI and AIA data, motivated the development of FlareBench.

The reason for using HMI and AIA in this dataset is that simultaneously using data from both the photospheric magnetic field and the multi-layered coronal atmosphere is crucial for effective solar flare prediction. HMI observes the line-of-sight and vector magnetic fields in the photosphere \cite{Schou_SoPy_2012}, whereas AIA captures extreme ultraviolet emissions from the multi-layered coronal atmosphere \cite{Lemen_SoPh_2012}. Solar flares are fundamentally linked to changes in magnetic field structure and reconnection \cite{Shibata_LivingReviews_2011}, highlighting the importance of incorporating data from both instruments.

\begin{figure*}[h]
    \centering
    \includegraphics[width=0.95\linewidth]{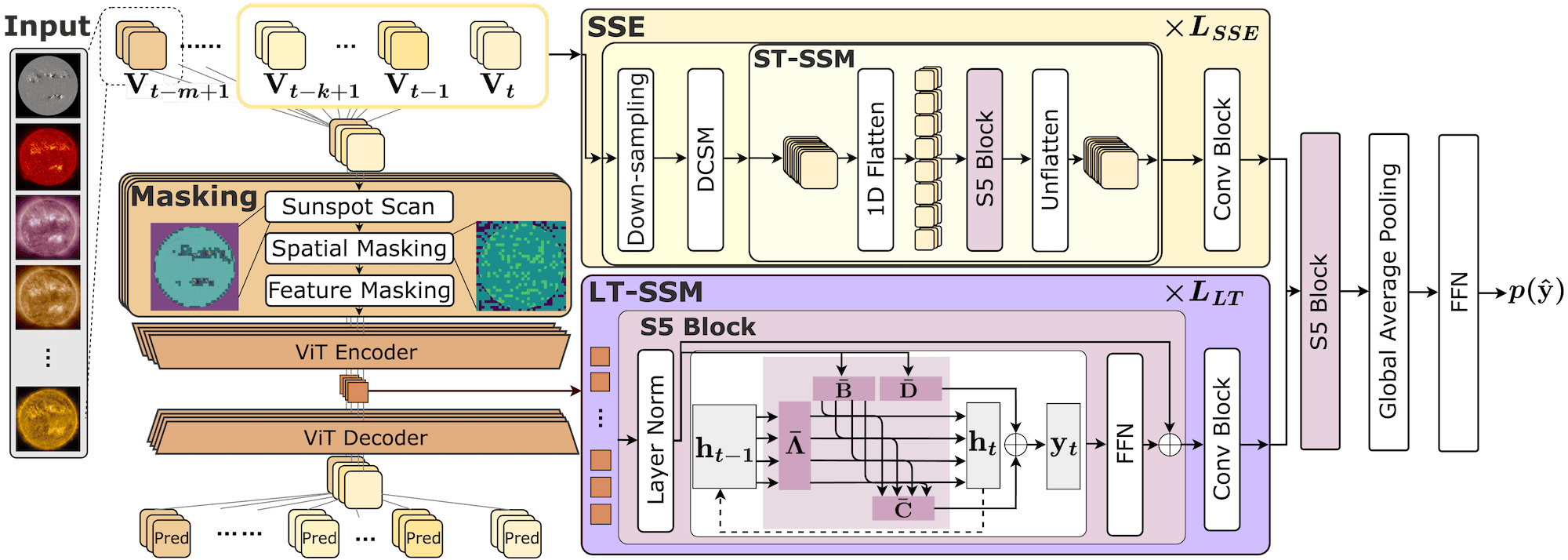}
    \vspace{-1mm}
    \caption{Architecture of the proposed method for solar flare prediction. The SSE extracts spatio-temporal features, incorporating a DCSM and a ST-SSM. The LT-SSM captures long-range temporal dependencies, employing multiple S5 Blocks. 
}    
\vspace{-2mm}
    \label{fig:model}
\end{figure*}

\vspace{-2mm}
\paragraph{Dataset statistics.}
\vspace{-1mm}
Fig.~\ref{fig:dist_aplit} shows the statistics of FlareBench.
The dataset initially comprised 100,801 samples collected from June 2011 to November 2022.
Each sample consists of a sequence of $k$ time steps. Each time step includes one HMI image and nine AIA images at different wavelengths, resulting in 10 channels. The input sequence is aligned at a 1-hour cadence. 
We excluded 2,440 samples with missing class labels and 2,524 samples where more than 25\% of the input channels (i.e., more than 25\% of the $k$ $C$-channels) were missing. 
Consequently, our dataset consisted of 95,837 samples, used for model training and evaluation. 
After these exclusions, the average missing rate of input channels across the remaining samples was 0.93\% (±4.03\%).
For these remaining samples, if less than 25\% of the $k$ $C$-channels were missing, we applied zero-padding to the missing channels.
The numbers of samples with ground truth labels for classes X, M, C, and O were 1,750, 13,263, 34,978, and 47,775, respectively.

Fig.~\ref{fig:dist_aplit} (b) illustrates the distribution of solar flare classes.
This distribution is highly imbalanced and varies significantly across different years. 
For example, no X-class flares occurred between 2018 and 2020, whereas 6.0\% of the flares in 2014 were X-class.
This imbalance results from the Sun's 11-year activity cycle \cite{Gnevyshev_SolarActivityCycle_1967, Hathaway_SolarActivityCycle_2010}, which causes fluctuations in solar flare occurrence. 


\vspace{-2mm}
\paragraph{Time-series cross-validation.}
\vspace{-2mm}
We divided the dataset into training, validation, and test sets using time-series cross-validation \cite{Tashman_timeseries_2000}, as illustrated in Fig.~\ref{fig:dist_aplit} (a). The numbers within each bar represent the time duration in years of the corresponding split. Fig.~\ref{fig:dist_aplit} (b) illustrates the distribution of flare classes within the test set of each fold. The three folds are designed to cover representative phases of the approximately 11-year solar activity cycle. 
We used observations from 2021 and 2022 to ensure each fold's test set represents a distinct phase of the solar cycle. Thus, we can evaluate model performance across varied solar conditions.
Each fold's training set also spans multiple years to maximize the inclusion of various solar activity phases in the training data. As illustrated in Fig.~\ref{fig:dist_aplit} (b), X and M-class flares are scarce in 2020 and 2021. 
The validation set was strategically defined to encompass the initial period of the dataset, ensuring these crucial classes are included while maintaining the chronological order of the test set within each fold (as depicted in Fig.~\ref{fig:dist_aplit} (a)).
Details on the data sources and composition of FlareBench can be found in Appendix \ref{supp:flarebench}.

%% file: pages/section_4.tex
\vspace{-1mm}
\section{Methodology}
\label{seq:met}
\vspace{-1mm}

This study proposes Deep Space Weather Model, a novel architecture extending deep state space models for classifying the maximum solar flare class within a 24-hour horizon, utilizing multi-wavelength images.
Fig.~\ref{fig:model} shows the structure of the proposed method.
Our proposed method includes three main components: the Solar Spatial Encoder (SSE), the Long-range Temporal State Space Model (LT-SSM), and the Sparse MAE as a pretraining model. The SSE incorporates the Depth-wise Channel Selective Module (DCSM) and the Spatio-Temporal State Space Model (ST-SSM).

The input to our model, denoted as $\mathbf{x}\in \mathbb{R}^{k \times C \times H \times W}$, is defined as $\mathbf{x} = (\mathbf{V}_{t-k+1}, \mathbf{V}_{t-k+2}, ..., \mathbf{V}_{t})$, where $\mathbf{V}_{t} \in \mathbb{R}^{C \times H \times W}$ represents a $C$-channel image at time step $t$. Here, $k$, $H$, and $W$ denote the history length, the height, and the width of the image, respectively. Each channel within $\mathbf{V}_{t}$ corresponds to a specific wavelength. 
Details on our multi-channel approach can be found in Appendix~\ref{supp:multi_channel_representation}.


\subsection{Solar Spatial Encoder}
\label{sec:met-sol}
\vspace{-1mm}
The SSE efficiently captures spatio-temporal features from
$\mathbf{x} \in \mathbb{R}^{k \times C \times H \times W}$,
and outputs
$\mathbf{h}_\mathrm{sse} \in \mathbb{R}^{L \times D}$,
where $L$ and $D$ represent the length of the encoded sequence and the feature dimension, respectively.
Unlike existing methods for solar flare prediction \cite{Park_ApJ_2018, Bhattacharjee_ApJ_2020, Li_AAS_2020}, which predominantly focus on limited temporal histories or fewer channels, the SSE can accommodate a broader range of spatio-temporal scales and an increased number of channels without incurring prohibitive computational costs. This flexibility is crucial for capturing subtle precursor patterns and for modeling complex solar phenomena that may span varied time windows. 

The SSE is composed of a multi-stage hierarchical architecture. 
The process begins with a 3D convolutional downsampling layer applied to $\mathbf{x}$, reducing the spatial resolution and producing $\mathbf{h_\mathrm{sse}}^{(0)} \in \mathbb{R}^{D \times C \times H^{(0)} \times W^{(0)}}$. Subsequently, the model iterates through $L_\mathrm{SSE}$ stages. In each stage $l$ ($1 \le l \le L_\mathrm{SSE}$), the following operations are performed sequentially:

\begin{enumerate}[label=\arabic*., leftmargin=*, itemsep=0.2em, topsep=0.2em]
    \item A 3D convolution downsamples the spatial resolution of $\mathbf{h_\mathrm{sse}}^{(l-1)}$, resulting in $\mathbf{h}_\mathrm{ds}^{(l)}$.

    \item The DCSM refines $\mathbf{h}_\mathrm{ds}^{(l)}$ using parallel 2D and 3D convolutions and channel weighting, producing $\mathbf{h}_\mathrm{dcs}^{(l)}$.

    \item The ST-SSM captures long-range dependencies in $\mathbf{h}_\mathrm{dcs}^{(l)}$, producing $\mathbf{h_\mathrm{sse}}^{(l)}$.
\end{enumerate}



After the final stage, $\mathbf{h_\mathrm{sse}}^{(L_\mathrm{SSE})}$ is processed through a sequence of 2D convolutional layers and then flattened, resulting in the final output $\mathbf{h}_\mathrm{sse} \in \mathbb{R}^{L \times D}$.

\subsection{Depth-wise Channel Selective Module}
\vspace{-1mm}
\label{seq:depth-wise}
The DCSM is designed to process each channel independently and selectively weight their importance. In multi-wavelength solar imaging, each channel represents different physical processes occurring at various atmospheric layers.
When all channels are treated equally, critical cues related to flaring events may be overshadowed, potentially degrading prediction performance. 

To address this problem, the DCSM first extracts spatio-temporal features through parallel 2D and 3D convolutions and then applies a channel-wise weighting mechanism to highlight more informative channels.
Given $\mathbf{h}_{\mathrm{ds}}^{(l)} \in \mathbb{R}^{D \times C^{(l)} \times H^{(l)} \times W^{(l)}}$ from the previous downsampling stage, the DCSM outputs $\mathbf{h}_{\mathrm{dcs}}^{(l)} \in \mathbb{R}^{D \times C^{(l)} \times H^{(l)} \times W^{(l)}}$ with identical dimensions. 



The DCSM employs parallel 2D and 3D convolutions to extract spatial and spatio-temporal features from $\mathbf{h}_{\mathrm{ds}}^{(l)}$, resulting in a fused feature map $\mathbf{F}_{\mathrm{fused}}$. Further details can be found in Appendix \ref{supp:sub-parallel-conv}.

A channel-wise weighting mechanism then emphasizes the most relevant channels. First, average pooling reduces the spatial dimensions of $\mathbf{F}_{\mathrm{fused}}$ to $1\times 1$, creating a channel-wise descriptor. This descriptor is then passed through 3D convolution layers, followed by a sigmoid function, resulting in the channel-wise weight $\mathbf{W}$.  Using this, the output is computed as $ \mathbf{h}_{\mathrm{dcs}}^{(l)} = \mathrm{Conv}_{1\times1}(\mathbf{F}_{\mathrm{fused}} \odot \mathbf{W}) + \mathbf{h}_{\mathrm{ds}}^{(l)} $, where $\odot$ and $\mathrm{Conv}_{1\times1}$ denote the Hadamard product and a $1\times 1$ convolution, respectively. 
\begin{figure}[t]
    \centering
    \includegraphics[width=\linewidth]{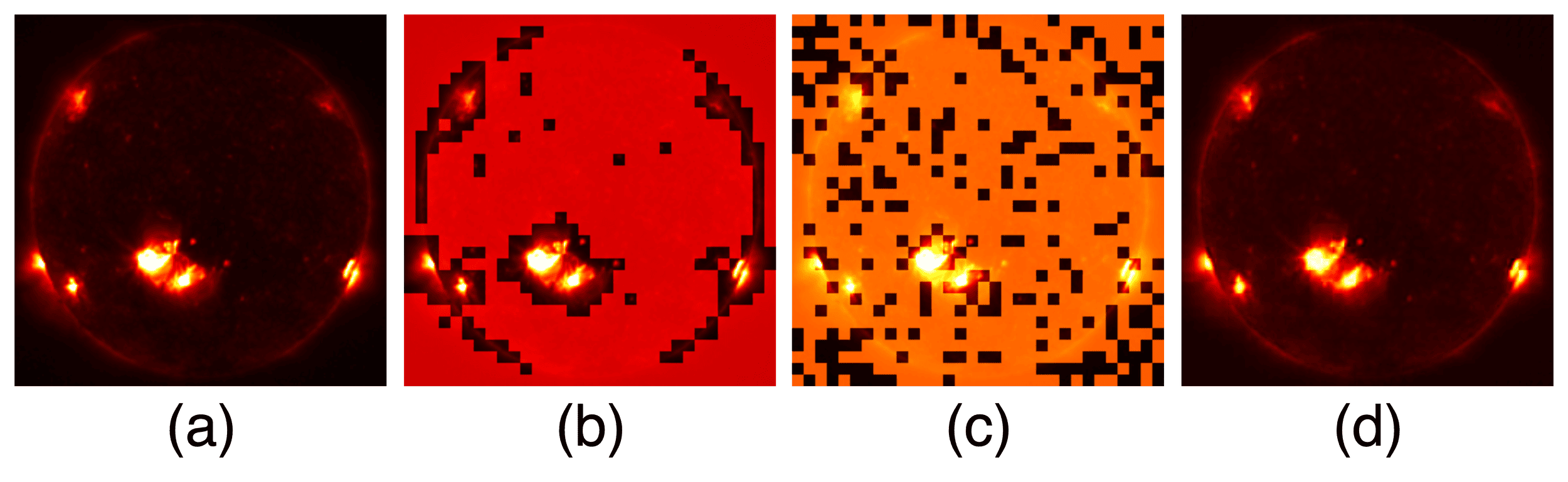}
    \vspace{-8mm}
    \caption{(a) An example of an original image, $V_t$. (b) Visualization of patches with the top $\alpha$\% highest standard deviation highlighted, often corresponding to sunspot regions. (c) Illustration of the spatial-level masking. (d) Reconstructed image by our proposed Sparse MAE.
}
    \label{fig:pretraining}
    \vspace{-3mm}
\end{figure}

\vspace{-1mm}
\subsection{Spatio-Temporal State Space Model}
\vspace{-1mm}
\label{seq:spatio-temp}
The ST-SSM aims to capture long-range dependencies in both spatial and temporal dimensions of multi-channel, long time-series solar images, leveraging the efficiency and effectiveness of deep state space models (deep SSMs). Specifically, we adopt S5 \cite{Smith_S5_2023} owing to its time-invariant, MIMO structure, which is particularly well-suited for efficiently modeling multi-channel, continuous solar image modalities. 
A detailed justification for adopting S5 is provided in Appendix \ref{supp:sub-justification}.
The ST-SSM maps the output of the DCSM, denoted as \(
\mathbf{h}_{\mathrm{dcs}}^{(l)} \in \mathbb{R}^{D \times C^{(l)} \times H^{(l)} \times W^{(l)}},
\)
to
\(
\mathbf{h}_{\mathrm{st}}^{(l)} \in \mathbb{R}^{D \times C^{(l)} \times H^{(l)} \times W^{(l)}}.
\)

Hereafter, \(\mathrm{SSM}(\cdot)\) refers to this S5-based operation described in Appendix \ref{supp:sub-s5-math}. 
An SSM block, denoted as \(\mathrm{SSMBlock}(\cdot)\), is introduced to provide a function similar to the Transformer's feedforward network.
For an input \(\mathbf{z}\), the \(\mathrm{SSMBlock}(\cdot)\) is computed as follows:
\begin{align}
\mathbf{z}' &= \mathrm{SSM}\bigl(\mathrm{LN}(\mathbf{z})\bigr) + \mathbf{z},\\
\mathrm{SSMBlock}(\mathbf{z}) &= \mathrm{MLP}\bigl(\mathrm{LN}(\mathbf{z}')\bigr) + \mathbf{z}',
\label{eq:ssmblock}
\end{align}
where \(\mathrm{LN}\), and \(\mathrm{MLP}\) represent a layer normalization operation, and a multi-layer perceptron, respectively.

Given \(\mathbf{h}_{\mathrm{dcs}}^{(l)}\), the ST-SSM first flattens and transposes the spatial and channel dimensions, creating an intermediate representation denoted as \(\bar{\mathbf{h}}_{\mathrm{st}}^{(l)} \in \mathbb{R}^{\bigl(C^{(l)} \cdot H^{(l)} \cdot W^{(l)}\bigr) \times D}\) by applying the transpose operation to the flattened \(\mathbf{h}_{\mathrm{dcs}}^{(l)}\).
The \(\mathrm{SSMBlock}\) is then applied to \(\bar{\mathbf{h}}_{\mathrm{st}}^{(l)}\), resulting in \(\tilde{\mathbf{h}}_{\mathrm{st}}^{(l)}\).
Finally, \(\tilde{\mathbf{h}}_{\mathrm{st}}^{(l)}\) is transposed and reshaped to \(\mathbf{h}_{\mathrm{st}}^{(l)} \in \mathbb{R}^{D \times C^{(l)} \times H^{(l)} \times W^{(l)}}\) so that essential spatio-temporal structures are preserved across channels.

\input{tab/quantitative_results}
\vspace{-1mm}
\subsection{Sparse MAE}
\label{sec:met-smae}
\vspace{-1mm}
We present Sparse MAE, a pretraining method for capturing long-range temporal dependencies in images with sparse but essential information regions.
Sparse MAE extends the MAE and tailors it specifically for the unique characteristics of solar images.
While MAEs have been successfully applied to video representation learning and other domains \cite{Reed_ICCV_2023, Cheng_ICCV_2023, Yang_CVPR_2023,Yin_CVPR_2024, Chen_ICCV_2023, Wang_CVPR_2023, Pei_CVPR_2024}, compressing spatial information, directly applying them to solar images presents unique challenges. For instance, crucial information for solar flare prediction, such as that from sunspot regions, can be completely masked, making it difficult to reconstruct these regions based solely on surrounding information if they are heavily masked. 
Sparse MAE addresses this challenge with a two-phase masking strategy during pretraining (Fig.~\ref{fig:pretraining}). 

For pretraining, the input, denoted as $\mathbf{x}_\mathrm{pre}$, is defined as $\mathbf{x}_\mathrm{pre} = (\mathbf{V}_{t-m+1}, \mathbf{V}_{t-m+2}, \dots, \mathbf{V}_{t}) \in \mathbb{R}^{m \times C \times H \times W}$, where $m$ ($> k$) represents the longer history length used in the pretraining. For each time step from $t-m+1$ to $t$, the model processes each $\mathbf{V}_t$ within the $\mathbf{x}_\mathrm{pre}$ independently.
The two-phase masking strategy proceeds as follows.

\vspace{-3mm}
\paragraph{Spatial-level masking.} As depicted in Figs.~\ref{fig:pretraining} (a) and (b), we first identify patches with high standard deviation (top $\alpha$\%). These patches often correspond to sunspot regions, critical for solar flare prediction. These high-variance patches are masked with a lower ratio, $r_l$, and the remaining patches are masked with a higher ratio, $r_h$ ($> r_l$). This preferential masking ensures that crucial features are less likely to be completely obscured. Fig.~\ref{fig:pretraining} (c) shows this masking.

\vspace{-3mm}
\paragraph{Feature-level masking.} After spatial masking, the resulting features are further masked with a ratio, $r_f$. This second masking phase prevents the MAE from relying solely on the easily reconstructed unmasked regions. This phase is critical when spatially sparse, yet vital, information (like sunspots) might be completely masked in the first phase. 
\vspace{1mm}

The encoder and decoder architectures, as well as the reconstruction loss, are identical to those in MAE \cite{He_MAE_2022}. Details are provided in the Appendix \ref{supp:setup}.
We repeat this process for each time step, resulting in a sequence of intermediate feature representations $\mathbf{h}_\mathrm{pre} = (\mathbf{h}_{t-m+1}, \dots, \mathbf{h}_{t-1}, \mathbf{h}_{t})$ used in subsequent modules. Each $\mathbf{h}_t$ is a $D_\mathrm{pre}$-dimensional feature vector encoding information from the corresponding time step.

\vspace{-1mm}
\subsection{Long-Range Temporal State Space Model}
\label{sec:met-lt}
\vspace{-1mm}
The LT-SSM is designed to accurately capture long-range temporal dependencies, spanning the solar rotation period, within the intermediate features obtained from the pretraining stage applied to solar images.
The LT-SSM extends deep SSMs to efficiently capture and model these long-range temporal relationships within the intermediate features $\mathbf{h}_\mathrm{pre}$, and it outputs $\mathbf{h}_\mathrm{lt} \in \mathbb{R}^{L \times D}$.

The LT-SSM comprises a series of $L_\mathrm{LT}$ SSM Blocks (defined in Equation~(\ref{eq:ssmblock})). The $l$-th SSM Block, where $1 \le l \le L_\mathrm{LT}$, outputs $\mathbf{h}_\mathrm{lt}^{(l)}$. The first SSM Block's output is obtained by applying the $\mathrm{SSMBlock}$ to $\mathbf{h}_\mathrm{pre}$. For subsequent layers, each SSM Block's output is obtained by applying the $\mathrm{SSMBlock}$ to the preceding SSM Block's output. Finally, the output $\mathbf{h}_\mathrm{lt}^{(L_\mathrm{LT})}$ is processed by 1D convolutional layers that adjust the channel dimension and produce the final output $\mathbf{h}_\mathrm{lt}$.

The outputs $\mathbf{h}_\mathrm{sse}$ and $\mathbf{h}_\mathrm{lt}$ from the SSE and LT-SSM, respectively, are integrated along the sequential dimension. We then obtain the predicted probability $p(\hat{\mathbf{y}})$ of the solar flare class corresponding to $\mathbf{x}$ as follows:
\vspace{-2mm}
\begin{equation}
p(\hat{\mathbf{y}}) = \text{FFN}(\text{SSMBlock}([\mathbf{h}_\mathrm{sse}; \mathbf{h}_\mathrm{lt}])),
\label{eq:prediction}
\vspace{-2mm}
\end{equation}
where SSMBlock and FFN represent an SSMBlock as defined in Eq. ~(\ref{eq:ssmblock}) and a feedforward network, respectively.

Our loss function comprises the cross-entropy loss, along with the GMGS and BSS losses proposed by \cite{Kaneda_FlareTransformer_2022}. Definitions of the loss function components, including their mathematical formulations, can be found in Appendix \ref{supp:sub-loss}.

\begin{figure*}[t]
    \centering
    \includegraphics[width=0.9\linewidth]{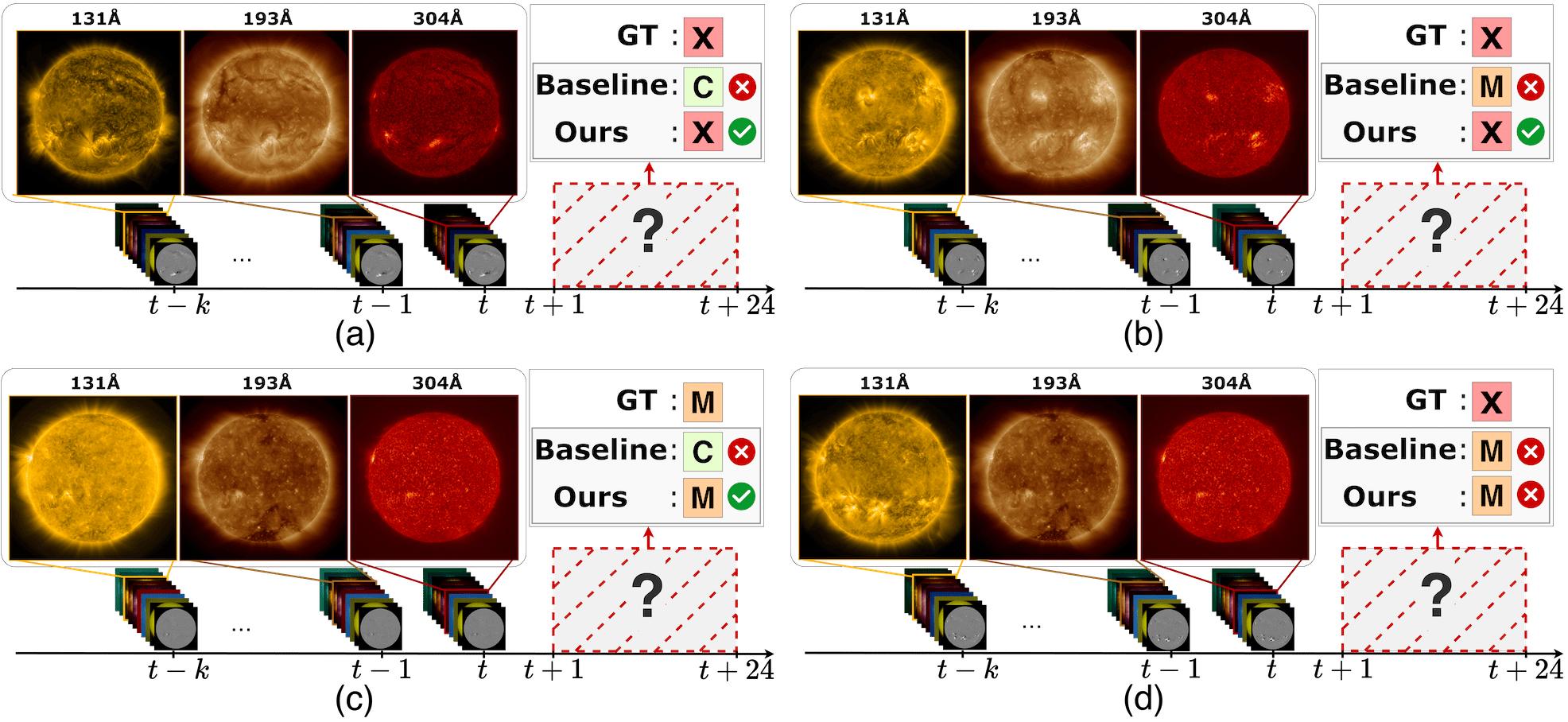}
    \vspace{-3mm}
    \caption{Qualitative results for flare predictions. (a) and (b) illustrate successful X-class predictions, (c) illustrates a successful M-class prediction, and (d) illustrates a failed X-class prediction. For each case, we present Extreme Ultraviolet images at 131 Å, 193 Å, and 304 Å from time $t-k$ to $t$, the ground truth (GT) label, and the predictions from the baseline and our proposed method.}
    \label{fig:qualitative_results}
    \vspace{-3mm}
\end{figure*}

%% file: tab/quantitative_results.tex
\begin{table*}[t]
\centering
\normalsize
\setlength{\tabcolsep}{4pt} 
\renewcommand{\arraystretch}{1.2}
\scalebox{0.95}{%
\begin{tabular}{@{}l l c c c @{}} 
 \toprule
 Method & \multicolumn{1}{c}{Test period} & GMGS$\uparrow$ & BSS$_{\geq \text{M}}$$\uparrow$ & TSS$_{\geq \text{M}}$$\uparrow$ \\
 \midrule\midrule
 Flare Transformer \cite{Kaneda_FlareTransformer_2022} (w/o PF) & 2014-2017 (4 years) & 0.220\footnotesize{ ± 0.116} & -1.770\footnotesize{ ± 0.225} & 0.198\footnotesize{ ± 0.371} \\
 DeFN-R \cite{Nishizuka_bss_2020} & 2014-2015 (2 years) & 0.302\footnotesize{ ± 0.055} & 0.036\footnotesize{ ± 0.982} & 0.279\footnotesize{ ± 0.162} \\
  CNN-LSTM & 2019-12-01 - 2022-11-30 (3 years) & 0.315\footnotesize{ ± 0.166} & 0.272\footnotesize{ ± 0.259} & 0.330\footnotesize{ ± 0.306} \\
 DeFN \cite{Nishizuka_DeFN_2018} & 2014-2015 (2 years) & 0.375\footnotesize{ ± 0.141} & 0.022\footnotesize{ ± 0.782} & 0.413\footnotesize{ ± 0.150} \\
 Flare Transformer \cite{Kaneda_FlareTransformer_2022} (full) & 2014-2017 (4 years) & 0.503\footnotesize{ ± 0.059} & 0.082\footnotesize{ ± 0.974} & 0.530\footnotesize{ ± 0.112} \\
 \rowcolor{red!10}
 Ours & 2019-12-01 - 2022-11-30 (3 years) & \textbf{0.582}\footnotesize{\textbf{ ± 0.032}} & \textbf{0.334}\footnotesize{\textbf{ ± 0.299}} & \textbf{0.543}\footnotesize{\textbf{ ± 0.074}} \\
 \noalign{\hrule height \arrayrulewidth}
 
  \begin{tabular}[b]{@{}l@{}}Human experts \cite{Kubo_humanperf_2017, Murray_humanperf_2017}\end{tabular} & 2000-2015 (16 years) & 0.48 & 0.16 & 0.50 \\

 \bottomrule
\end{tabular}
}
\vspace{-2.3mm}
\caption{Quantitative comparison. The best scores are in bold. PF stands for physical features.}
\vspace{-3mm}
\label{tab:quantitative_results}
\end{table*}

%% file: pages/section_5.tex
\vspace{-1mm}
\section{Experiments}
\label{seq:exp}

%% file: pages/section_6.tex
\subsection{Setup}

\vspace{-1mm}
\paragraph{Baselines.}
Several approaches have been successfully applied to solar flare prediction, including those based on MLPs, CNNs or LSTMs, and Transformers. 
We selected representative methods as baseline methods for comparison.
For MLP-based methods, we include DeFN \cite{Nishizuka_DeFN_2018} and DeFN-R \cite{Nishizuka_bss_2020}. 
Among CNN-based or LSTM-based methods, we adopted a CNN-LSTM similar to \cite{Sun_AAS_2022}, which is designed to process multi-channel images. For Transformer-based methods, we selected Flare Transformer \cite{Kaneda_FlareTransformer_2022}, which leverages both images and physical features.

We used their previously reported results for DeFN, DeFN-R, and Flare Transformer because these methods rely on physical features that are not fully provided in FlareBench.
For CNN-LSTM, we reproduced and evaluated it on FlareBench under the same conditions as our proposed method.

\vspace{-3mm}
\paragraph{Evaluation metrics.}
\label{sec:exp-eva}
For evaluating performance in the context of imbalanced solar flare observations, we employ three standard metrics given their suitability to this task\cite{Kubo_humanperf_2017, Nishizuka_bss_2020, Kaneda_FlareTransformer_2022,  Abduallah_SolarFlareNet_2023, Li_AAS_2024}: GMGS\cite{Gandin_gmgs_1992}, BSS$_{\scalebox{0.8}{$\geq \mathrm{M}$}}$\cite{Nishizuka_bss_2020}, and TSS$_{\scalebox{0.8}{$\geq \mathrm{M}$}}$\cite{Kubo_humanperf_2017}.
The significant class imbalance in solar flare prediction (detailed in Subsection~\ref{seq:flarebench}) necessitates careful metric selection. Maximizing accuracy alone can be misleading, as a naive model predicting the majority class (e.g., ``C-class or O class'') can achieve high accuracy without capturing the underlying patterns of flare occurrence. 

GMGS ensures fair evaluation across all flare classes \cite{Gandin_gmgs_1992}. BSS$_{\scalebox{0.8}{$\geq \mathrm{M}$}}$ assesses the forecast reliability for larger ($\geq$M) and smaller ($<$M) flares \cite{Nishizuka_bss_2020}. TSS$_{\scalebox{0.8}{$\geq \mathrm{M}$}}$ balances the accurate prediction of both larger and smaller flares \cite{Kubo_humanperf_2017}. For these metrics, $\geq$M and $<$M indicate that the evaluation is performed by categorizing flares as (a) M-class or above or (b) below M-class, respectively.
Detailed definitions of these metrics, including the mathematical formulations for GMGS, BSS, and TSS, can be found in Appendix \ref{supp:setup}. Additional experimental setup details can also be found there.

\subsection{Quantitative Results}
\vspace{-1mm}
Table \ref{tab:quantitative_results} shows the quantitative comparison between our proposed method and the baseline methods. The values represent the mean and standard deviation of each metric, calculated using time-series cross-validation. The corresponding test set periods for each method are provided in the table.
The table also presents the performance of human experts, which was reported by Kubo et al. for daily forecasting operations spanning 2000 to 2015 \cite{Kubo_humanperf_2017}.
Details on the human forecasters can be found in Appendix \ref{supp:human_forecasters}.

Table \ref{tab:quantitative_results} shows that our method achieved a GMGS, BSS$_{\scalebox{0.8}{$\geq \mathrm{M}$}}$, and TSS$_{\scalebox{0.8}{$\geq \mathrm{M}$}}$ of 0.582, 0.334, and 0.543, respectively. 
Our method outperformed the CNN-LSTM by 0.267, 0.062, and 0.213 in GMGS, BSS$_{\scalebox{0.8}{$\geq \mathrm{M}$}}$, and TSS$_{\scalebox{0.8}{$\geq \mathrm{M}$}}$, respectively. 
Furthermore, our method also surpassed Flare Transformer \cite{Kaneda_FlareTransformer_2022}, which had the highest scores among the baselines in all metrics, by 0.079, 0.252, and 0.013 in GMGS, BSS$_{\scalebox{0.8}{$\geq \mathrm{M}$}}$, and TSS$_{\scalebox{0.8}{$\geq \mathrm{M}$}}$, respectively. 
Notably, our method even outperformed human expert performance by margins of 0.102, 0.174, and 0.043 in GMGS, BSS$_{\scalebox{0.8}{$\geq \mathrm{M}$}}$, and TSS$_{\scalebox{0.8}{$\geq \mathrm{M}$}}$, respectively.
Our method demonstrated statistically significant differences in GMGS compared with the CNN-LSTM ($p < 0.05$). 
However, a direct statistical comparison with other baselines was not feasible because of differences in their test set periods.

It is particularly noteworthy that the results surpass the performance of human experts across three distinct metrics, utilizing the dataset covering a complete solar cycle.

\begin{figure}[t]
    \centering
    \includegraphics[width=0.95\linewidth]{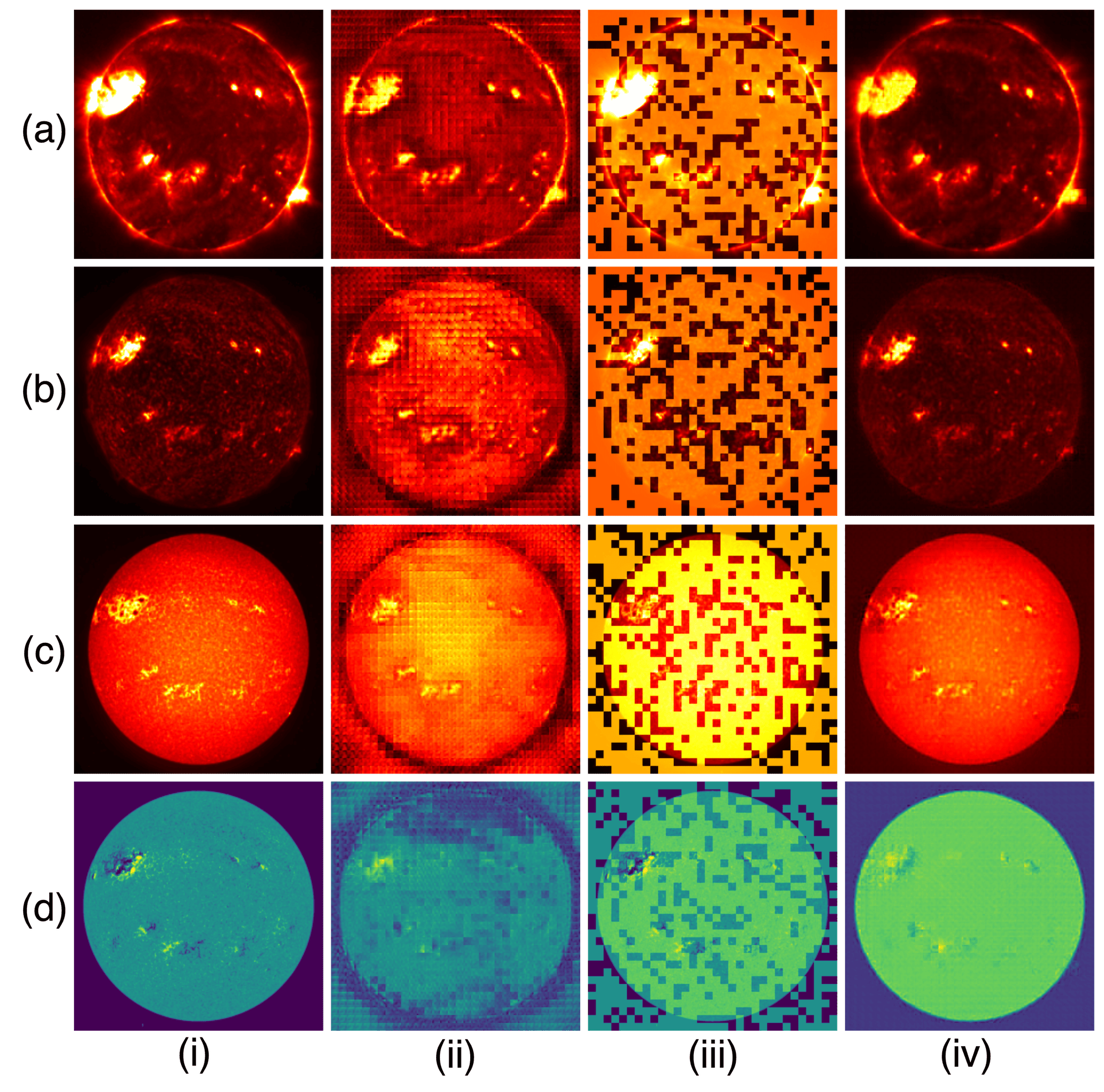}
    \vspace{-2mm}
    \caption{Reconstruction results obtained from the MAE \cite{He_MAE_2022} with a mask ratio ($\rho$) of 0.5 and Sparse MAE. Rows (a), (b), (c), and (d) include 94 Å AIA, 304 Å AIA, 1600 Å AIA, and HMI images, respectively, captured three hours before an upcoming X-class flare. Columns (i), (ii), (iii), and (iv) present the original image, the baseline reconstruction, the spatial-level masking of the Sparse MAE, and the Sparse MAE reconstruction, respectively.}
    \label{fig:qualitative_results_pretraining}
    \vspace{-3mm}
\end{figure}

\subsection{Qualitative Results}
\vspace{-1mm}

Fig.~\ref{fig:qualitative_results} shows qualitative results of our method for both X-class and M-class flare predictions. In the figure, we show Extreme Ultraviolet images at 131, 193, and 304 Å from time $t-k$ to $t$, the ground truth label, and the predictions from both the baseline CNN-LSTM and our method.

Subfigures (a) and (b) illustrate successful X-class predictions, (c) illustrates a successful M-class prediction, while the baseline CNN-LSTM misclassified the flare class.

Subfigure (d) illustrates a failed X-class prediction.
Both our model and the baseline CNN-LSTM incorrectly predicted an M-class flare. This misclassification may be due to the challenging nature of this sample. 
During the 24 hours preceding the X-class flare, there were two distinct peaks in X-ray flux, each corresponding to M-class flares.
Moreover, the X-class flare lay at the boundary between X- and M-class events.
These factors likely contributed to the difficulty in accurately classifying the flare class. The proximity of the observed X-class event to the M/X-class boundary and the preceding M-class activity suggests that this was a challenging boundary case. The X-ray flux transitions for this sample can be found in Appendix \ref{supp:xray_transitions}.



\vspace{-1mm}
\subsection{Qualitative Results for Pretraining}
\vspace{-1mm}
\label{seq:qualitative_pretraining}
Fig.~\ref{fig:qualitative_results_pretraining} illustrates the reconstruction results obtained from the baseline MAE \cite{He_MAE_2022} with a mask ratio ($\rho$) of 0.5 and our proposed Sparse MAE. Rows (a), (b), (c), and (d) illustrate 94 Å AIA images, 304 Å AIA images, 1600 Å AIA images, and HMI images, respectively, captured three hours before an upcoming X-class flare. Columns (i), (ii), (iii), and (iv) present the original image, the baseline reconstruction, the spatial-level masking of the Sparse MAE, and the Sparse MAE reconstruction, respectively.

As depicted in subfigures (a-ii) and (a-iv), and the others, the Sparse MAE reconstructs features in and around sunspots with high fidelity. By contrast,  the baseline method struggles to reproduce fine details in these regions.
These observations suggest that the enhanced representation of sunspots in the Sparse MAE reconstructions can be attributed to its two-phase masking strategy, which emphasizes preserving essential features such as sunspots. Further results can be found in Appendix \ref{supp:qualitative_pretraining}.

\input{tab/ablation_module}

\input{tab/ablation_pretraining}
\vspace{-1mm}
\subsection{Ablation Study}
\vspace{-1mm}
We conducted ablation studies to demonstrate the effectiveness of our proposed method.  We performed module-wise ablation and pretraining ablation. Details on deep SSM ablation can be found in Appendix \ref{supp:ablation_study}.

\vspace{-2mm}
\paragraph{Module-wise ablation.}
Table \ref{tab:ablation_module} presents the performance impact of three main modules: LT-SSM, DCSM, and ST-SSM. We compare five model configurations: (1-i) exclusion of LT-SSM, (1-ii) exclusion of DCSM, (1-iii) exclusion of ST-SSM, (1-iv) exclusion of the entire SSE (both DCSM and ST-SSM), and (1-v) the complete model including all three modules. Here, (D), (S), and (L) represent DCSM, ST-SSM, and LT-SSM, respectively.
Regarding GMGS, Models (1-i), (1-ii), (1-iii) and (1-iv) exhibited lower performance compared with Model (1-v) by 0.252, 0.240, 0.211 and 0.088 points, respectively. These results indicate that each module improves model performance, with the LT-SSM module exerting the most significant impact.

Maximizing BSS$_{\scalebox{0.8}{$\geq \mathrm{M}$}}$ alone can be misleading because of the class imbalance in the solar flare class distribution (where O and C classes are prevalent). It is possible to achieve high BSS$_{\scalebox{0.8}{$\geq \mathrm{M}$}}$ by primarily predicting the majority classes while performing poorly on the critical but rarer M- and X-class flares. Therefore, it is crucial to improve GMGS (TSS$_{\scalebox{0.8}{$\geq \mathrm{M}$}}$) along with BSS$_{\scalebox{0.8}{$\geq \mathrm{M}$}}$, achieving a balance that reflects strong performance on both frequent and rare, high-impact events.

\vspace{-2mm}
\paragraph{Pretraining ablation.}
Table \ref{tab:ablation_pretraining} presents an ablation study comparing the impact of different pretraining methods on solar flare prediction performance. We evaluate three models: (2-i) using MAE \cite{He_MAE_2022} with a mask ratio ($\rho$) of 0.75, (2-ii) using MAE \cite{He_MAE_2022} with $\rho$ = 0.5, and (2-iii) our proposed Sparse MAE.
The table presents the GMGS, BSS$_{\scalebox{0.8}{$\geq \mathrm{M}$}}$, and TSS$_{\scalebox{0.8}{$\geq \mathrm{M}$}}$ scores for the solar flare prediction task, and the MSE from the pretraining phase. MSE are computed as the mean squared error over the masked patches. 

The results reveal differences in performance across the evaluated models.  Models (2-i) and (2-ii) underperform Model (2-iii) in terms of both GMGS and MSE. Specifically, Model (2-i) underperforms Model (2-iii) by 0.296 and 1.686 points in GMGS and MSE, respectively, while Model (2-ii) underperforms Model (2-iii) by 0.162 and 3.426 points in GMGS and MSE, respectively. These results indicate that the improved reconstruction of crucial solar regions, such as sunspots, achieved with Sparse MAE, leads to extracting features more relevant for this task.

%% file: tab/ablation_module.tex
\begin{table}[t]
  \centering
  \normalsize
  \setlength{\tabcolsep}{1.5pt}
  \renewcommand{\arraystretch}{1.2}
  \scalebox{0.85}{%
  \begin{tabular}{
    >{\centering\arraybackslash}p{0.06\textwidth} 
    >{\centering\arraybackslash}p{0.03\textwidth}
    >{\centering\arraybackslash}p{0.03\textwidth}
    >{\centering\arraybackslash}p{0.03\textwidth}
    >{\centering\arraybackslash}p{0.12\textwidth}
    >{\centering\arraybackslash}p{0.12\textwidth}
    >{\centering\arraybackslash}p{0.12\textwidth}
  }
    \toprule
    Model & (D) & (S) & (L) & GMGS$\uparrow$ & BSS$_{\geq\text{M}}$$\uparrow$ & TSS$_{\geq\text{M}}$$\uparrow$ \\
    \hhline{*{7}{=}}
    (1-i)   & \checkmark & \checkmark &  & $0.330 \pm \scriptstyle 0.029$ & $-0.084 \pm \scriptstyle 0.487$ & $0.381 \pm \scriptstyle 0.018$ \\
    (1-ii)  &            & \checkmark & \checkmark & $0.342 \pm \scriptstyle 0.180$ & $0.320 \pm \scriptstyle 0.052$ & $0.438 \pm \scriptstyle 0.165$ \\
    (1-iii) & \checkmark &            & \checkmark & $0.371 \pm \scriptstyle 0.094$ & $\mathbf{0.392 \pm \scriptstyle 0.053}$ & $0.413 \pm \scriptstyle 0.060$ \\
    (1-iv)  &            &            & \checkmark & $0.494 \pm \scriptstyle 0.129$ & $0.232 \pm \scriptstyle 0.630$ & $0.462 \pm \scriptstyle 0.175$ \\
    \rowcolor{red!10}
    (1-v)   & \checkmark & \checkmark & \checkmark & $\mathbf{0.582 \pm \scriptstyle 0.032}$ & $0.334 \pm \scriptstyle 0.299$ & $\mathbf{0.543 \pm \scriptstyle 0.074}$ \\
    \bottomrule
  \end{tabular}
  }
    \vspace{-2mm}
      \caption{Ablation study on the core modules in our method.}
    \label{tab:ablation_module}
\end{table}

%% file: tab/ablation_pretraining.tex
\begin{table}[t]
  \centering
  \normalsize
  \setlength{\tabcolsep}{4pt}
  \renewcommand{\arraystretch}{1.2}
  \vspace{-2mm}
  \scalebox{0.8}{%
    \begin{tabular}{l c c c c} 
      \toprule
      Model & GMGS$\uparrow$ & BSS$_{\geq\text{M}}$$\uparrow$ & TSS$_{\geq\text{M}}$$\uparrow$ & MSE$\downarrow$ \\
      \hhline{=====} 
      (2-i)  &
      $0.286 \pm \scriptstyle 0.090$ &
      $0.067 \pm \scriptstyle 0.306$ &
      $0.428 \pm \scriptstyle 0.173$ &
      $4.147 \pm \scriptstyle 0.378$ \\
      (2-ii) &
      $0.420 \pm \scriptstyle 0.062$ &
      $\mathbf{0.354 \pm \scriptstyle 0.163}$ &
      $0.402 \pm \scriptstyle 0.100$ &
      $5.887 \pm \scriptstyle 0.329$ \\
      \rowcolor{red!10}
      (2-iii) &
      $\mathbf{0.582 \pm \scriptstyle 0.032}$ &
      $0.334 \pm \scriptstyle 0.299$ &
      $\mathbf{0.543 \pm \scriptstyle 0.074}$ &
      $\mathbf{2.461 \pm \scriptstyle 0.115}$ \\
      \bottomrule
    \end{tabular}
  }
    \vspace{-2mm}
  \caption{Ablation study: impact of masking strategies.}
  \label{tab:ablation_pretraining}
  \vspace{-3mm}
\end{table}

%% file: pages/section_7.tex
\vspace{-2mm}
\section{Conclusion}
\vspace{-1mm}

In this study, we focused on predicting the maximum class of solar flare within the next 24 hours. 
We propose the Deep Space Weather Model, a novel method that extends deep state space models to effectively capture long-range spatio-temporal dependencies and represent crucial, yet sparse, informative regions in multi-wavelength solar images. 
Furthermore, we have constructed a new public benchmark for solar flare prediction, FlareBench, covering a full 11-year solar activity cycle to mitigate the risk of biased evaluations inherent in shorter-duration datasets. Finally, our method is the first to achieve superhuman performance on both the GMGS and BSS compared with human experts.

\vspace{-4mm}
\paragraph{Limitations.}
Currently, our model uses compressed solar images, which may limit its ability to capture fine-grained spatial details relevant for accurate prediction. In future work, we plan to adopt full-resolution solar images rather than compressed representations to enable finer-grained spatial modeling and enhance prediction performance.

\section*{Acknowledgment}
This work was partially supported by JSPS KAKENHI Grant Number 23K28168 and NEDO.

\vspace{-2mm}

%% file: pages/suppl.tex
\clearpage
\appendix
\setcounter{page}{1}
\maketitlesupplementary

\section{Additional Related Work}
\label{supp:related_work}
\paragraph{Datasets and benchmarks.}
Observations from space have significantly enhanced our understanding of astronomical phenomena and played a crucial role in advancing solar physics. 
For example, star tracking and localization have improved thanks to recent advances like Chin et al.'s event-based pipeline \cite{Chin_CVPR_Workshops_2019} and the StarNet dataset \cite{Felt_WACV_2024} for narrow-field star localization.
In fields like satellite pose estimation, datasets such as SPEED \cite{Kisantal_TAES_2020} and SwissCube \cite{Hu_CVPR_2021} address challenges such as scale variations and adverse illumination. Similarly, in remote sensing, datasets like DOTA \cite{Gui_CVPR_2018} and EarthNet2021 \cite{Mesa_CVPRW_2021} are used to study dynamic terrestrial processes. These examples highlight the importance of tailored benchmarks for developing and evaluating task-specific methods.

\paragraph{Solar flare prediction.}
Numerous methods have been proposed for solar flare prediction, including early approaches using Multi-layer perceptrons (MLPs) and more recent methods employing Convolutional Neural Networks (CNNs) and Recurrent Neural Networks, such as Long Short-Term Memory networks (LSTMs). DeFN incorporates 79 features extracted from sunspot images, including features related to coronal hot brightening and X-ray intensity trends specifically chosen for operational forecasting. Subsequently, Li et al.\cite{Li_AAS_2020} propose a CNN model trained on Spaceweather HMI Active Region Patch (SHARP) magnetograms to predict flares, leveraging the ability of CNNs to extract spatial features. Concurrently, \cite{Liu_ApJ_2019} develops an LSTM to capture the temporal evolution of active regions using both magnetic parameters and flare history, demonstrating the importance of temporal dynamics in flare prediction.
However, these traditional approaches, including MLPs, CNNs, and LSTMs, while demonstrating potential for solar flare prediction, primarily rely on heuristic physical features or often have limitations in capturing long-range spatio-temporal dependencies.

More recently, Transformer-based models have been explored. For example, SolarFlareNet \cite{Abduallah_SolarFlareNet_2023} utilizes a transformer-based framework to predict flares from time series of SHARP parameters, extending the prediction window to 72 hours.

Despite their strength in modeling long-range dependencies, the computational cost of Transformers scales quadratically with sequence length. This cost is a significant challenge when applying Transformers to the long, multi-channel, full-disk solar image time series considered in our work, where computational and memory demands can become prohibitive.

\paragraph{Masked autoencoders.}
The Masked Autoencoder (MAE) approach introduced by He et al.\cite{He_MAE_2022} has inspired a wide range of extensions and adaptations across various domains. 
Following the MAE, numerous extensions have explored diverse applications and architectures, geospatial representation learning \cite{Reed_ICCV_2023}, motion forecasting \cite{Cheng_ICCV_2023}, 3D point clouds \cite{Yang_CVPR_2023}, and facial landmark estimation \cite{Yin_CVPR_2024}. For instance, Traj-MAE \cite{Chen_ICCV_2023} adapts MAE for trajectory prediction in autonomous driving, using diverse masking strategies and a continual pre-training framework to capture social and temporal interactions. In the realm of video understanding, several MAE-based methods have been proposed \cite{Wu_CVPR_2023, Tong_NeurIPS_2022, Wang_CVPR_2023, Gupta_NeurIPS_2023}. VideoMAC \cite{Pei_CVPR_2024} addresses the resource-intensive nature of many Vision Transformer-based approaches by combining masked autoencoders with ConvNets, using a dual encoder architecture for inter-frame consistency.

\paragraph{Deep SSMs.}
Deep SSMs are founded on the Linear State-Space Layer \cite{Gu_NeurIPS_2021}, inspired by classical state space models in control theory \cite{Kalman_JBE_1960}. They achieve efficient sequence modeling by leveraging the HiPPO matrix \cite{Gu_NeurIPS_2020}, which enables the memorization of input sequences through optimal polynomial approximation. For example, S4 \cite{Gu_ICLR_2022} introduces a method for learning the HiPPO matrix. Building upon S4, S5 \cite{Smith_S5_2023} proposes a new state space layer that utilizes a single multi-input, multi-output SSM instead of S4's bank of single-input, single-output SSMs. Furthermore, S5 uses an efficient parallel scan for computation, removing the need for the convolutional approach used in S4 and its associated convolution kernel computation.

\input{tab/flare_class}
\section{Flare Class Definitions}

\label{supp:flare_class}

Table \ref{tab:flare_class} shows the standard classification of solar flares based on their peak X-ray flux, $I$, measured in the 1-8 Å wavelength range by the X-ray Sensor (XRS) on board the Geostationary Operational Environmental Satellites (GOES). This classification is widely used in solar physics and space weather forecasting.

Within the X-class, flares are further categorized by a linear scale.  An X1.0 flare corresponds to a peak flux of $10^{-4}$ W/m$^2$.  The number following the 'X' indicates a multiple of this base value. For example, an X2.0 flare has a peak flux of $2 \times 10^{-4}$ W/m$^2$, an X3.0 flare has a peak flux of $3 \times 10^{-4}$ W/m$^2$, and so on.  This same linear scaling applies within the M and C classes as well.

\section{Deep Space Weather Model}
\label{supp:method}

\subsection{Multi-channel Representation Beyond Conventional RGB}
\label{supp:multi_channel_representation}
Our approach is analogous to how the three channels of an RGB image represent colors within the visible spectrum. However, by incorporating HMI and AIA images, we extend the concept of multi-channel representation to higher dimensionality, encompassing a broader range of the electromagnetic spectrum.

\subsection{Parallel 2D and 3D Convolutions}
\label{supp:sub-parallel-conv}
The DCSM begins by applying two parallel convolutional operations to $\mathbf{h}_{\mathrm{ds}}^{(l)}$:
\vspace{-2mm}
\begin{equation}
\mathbf{F}_{\mathrm{fused}} = \mathrm{Conv3D}(\mathbf{h}_{\mathrm{ds}}^{(l)}) + \mathrm{Conv2D}(\mathbf{h}_{\mathrm{ds}}^{(l)}),
\label{eq:fused}
\vspace{-2mm}
\end{equation}
where $\text{Conv3D}$ and $\text{Conv2D}$ denote a 3D convolution and a 2D convolution applied independently to each frame, respectively. The 3D convolution is intended to capture spatio-temporal patterns across channels. By contrast, the 2D convolution focuses on spatial features within each channel (e.g., sunspot structures). Their outputs are summed element-wise to produce the fused feature map $\mathbf{F}_{\mathrm{fused}}$.

\subsection{Justification for Adopting S5}
\label{supp:sub-justification}
We adopt the S5 \cite{Smith_S5_2023} to model these multi-channel solar images accurately based on the following considerations:
\paragraph{Time-invariance for continuous modalities.} Time-variant SSMs, such as Mamba \cite{Gu_Mamba_2024}, introduce selection mechanisms that may degrade performance on continuous modalities like solar images, especially multi-wavelength observations. By contrast, time-invariant (LTI) SSMs are suggested to perform better on continuous signals \cite{Gu_Mamba_2024}. Supporting this, the experiment on YouTubeMix in \cite{Gu_Mamba_2024} indicates that LTI models, such as S5, may be more suitable when the input is a continuous modality.
\vspace{-4mm}
\paragraph{MIMO structure for multi-channel efficiency.} Single-input, single-output (SISO) setups, commonly used in previous deep SSM architectures (e.g., \cite{Gu_ICLR_2022, Gu_NeurIPS_2022}), cannot fully leverage the multi-channel nature of the input. In contrast to the SISO approach, the multi-input multi-output (MIMO) structure of S5 permits direct modeling of inter-channel dependencies and offers improved computational efficiency.

\subsection{S5 Mathematical Formulation}
\label{supp:sub-s5-math}
The core of the S5 layer is a MIMO SSM, which can be represented in continuous time. Drawing inspiration from continuous system equations, an input \(\mathbf{u}(t) \in \mathbb{R}^D\), a latent state \(\mathbf{x}(t) \in \mathbb{R}^P\), and an output \(\mathbf{y}(t) \in \mathbb{R}^D\) are considered. The general form of a continuous-time linear SSM can be defined as:
\begin{equation}
\frac{d\mathbf{x}(t)}{dt} = \mathbf{A}\mathbf{x}(t) + \mathbf{B}\mathbf{u}(t), \quad \mathbf{y}(t) = \mathbf{C}\mathbf{x}(t) + \mathbf{D}\mathbf{u}(t),
\end{equation}
where \(\mathbf{A} \in \mathbb{C}^{P \times P}\), \(\mathbf{B} \in \mathbb{C}^{P \times D}\), \(\mathbf{C} \in \mathbb{C}^{D \times P}\), and \(\mathbf{D} \in \mathbb{R}^{D \times D}\) denote the state matrix, input matrix, output matrix, and feedthrough matrix, respectively.
To enable efficient parallel computation, \(\mathbf{A}\) is diagonalized as \(\mathbf{A} = \mathbf{V} \boldsymbol{\Lambda} \mathbf{V}^{-1}\), where \(\boldsymbol{\Lambda} \in \mathbb{C}^{P \times P}\) is a diagonal matrix and \(\mathbf{V} \in \mathbb{C}^{P \times P}\) is the matrix of eigenvectors, enabling  us to rewrite the system dynamics as:
\begin{equation}
\frac{d\tilde{\mathbf{x}}(t)}{dt} = \boldsymbol{\Lambda} \tilde{\mathbf{x}}(t) + \tilde{\mathbf{B}}\mathbf{u}(t), \quad \mathbf{y}(t) = \tilde{\mathbf{C}}\tilde{\mathbf{x}}(t) + \mathbf{D}\mathbf{u}(t),
\end{equation}
where \(\tilde{\mathbf{x}}(t) = \mathbf{V}^{-1}\mathbf{x}(t)\), \(\tilde{\mathbf{B}} = \mathbf{V}^{-1}\mathbf{B}\), and \(\tilde{\mathbf{C}} = \mathbf{C}\mathbf{V}\).
This is the reparameterized system with a diagonal state matrix, which is crucial for the efficiency of S5.

This diagonalized system is then discretized using the Zero-Order Hold (ZOH) method with learnable timescale parameters \(\boldsymbol{\Delta} \in \mathbb{R}^P\). The ZOH method assumes that the input function remains constant over each interval defined by the timescale parameter. In practice, the feedthrough matrix \(\mathbf{D}\) is restricted to be diagonal. Thus, the S5 layer has the learnable parameters \(\tilde{\mathbf{B}} \in \mathbb{C}^{P \times D}\), \(\tilde{\mathbf{C}} \in \mathbb{C}^{D \times P}\), \(\text{diag}(\mathbf{D}) \in \mathbb{R}^D\), \(\text{diag}(\boldsymbol{\Lambda}) \in \mathbb{C}^P\), and the timescale parameters \(\boldsymbol{\Delta} \in \mathbb{R}^P\). The performance of S5 is sensitive to the initialization of \(\mathbf{A}\). It is initialized with a diagonalized HiPPO-N matrix.


\subsection{Loss Function}
\label{supp:sub-loss}
Our loss function comprises three components: the conventional cross-entropy loss ($L_\mathrm{CE}$), the BSS loss ($L_\mathrm{BSS}$), and the GMGS loss ($L_\mathrm{GMGS}$). We employ the BSS and GMGS losses, introduced by \cite{Kaneda_FlareTransformer_2022}. The BSS loss is used to optimize the BSS, a proper scoring rule that comprehensively evaluates probabilistic predictions. Furthermore, the BSS loss is differentiable, enabling efficient optimization through gradient-based methods. The GMGS loss is used to improve the GMGS using its own score matrix for the weights in the loss calculation. 

The BSS loss is defined as:
\begin{align}
L_\mathrm{BSS} &= \frac{1}{N} \sum_{n=1}^{N} \sum_{i=1}^{I} \Bigl( p(\hat{y}_{ni}) - y_{ni} \Bigr)^2, \\
\intertext{and the GMGS loss is defined as:}
L_\mathrm{GMGS} &= -\frac{1}{N} \sum_{n=1}^{N} s_{i^*j^*} \sum_{i=1}^{I} y'_{ni} \log\Bigl( p(\hat{y}_{ni}) \Bigr),
\end{align}
where
\begin{align}
i^* &= \text{argmax}_i (y_{ni}), \\
j^* &= \text{argmax}_j (p(\hat{y}_{nj})),
\end{align}
$N$ and $I$ denote the number of samples and the number of classes, respectively. $y'_{ni}$ is the label-smoothed version of $y_{ni}$, $p(\hat{y}_{ni})$ is the predicted probability for the $i$-th class of the $n$-th sample, and $s_{i^*j^*}$ denotes the element from the GMGS score matrix corresponding to the true class $i^*$ and the predicted class $j^*$, respectively.

Our overall loss function is a weighted sum of these three components:
\begin{equation}
L = L_\mathrm{CE} + \lambda_\mathrm{GMGS} L_\mathrm{GMGS} + \lambda_\mathrm{BSS} L_\mathrm{BSS},
\end{equation}
where $\lambda_\mathrm{GMGS}$ and $\lambda_\mathrm{BSS}$ are the loss weights controlling the contributions of the GMGS and BSS losses, respectively.

\input{tab/experimental_setup}
\input{tab/experimental_setup_pretraining}

\section{FlareBench}
\label{supp:flarebench}
\paragraph{Data sources and composition.}
In this study, we constructed FlareBench by combining solar observation data from the Joint Space Operations Center (JSOC)\footnote{\url{http://jsoc.stanford.edu}.} with X-ray flux measurements from the Geostationary Operational Environmental Satellites. 
Our dataset includes:
\begin{enumerate}[label=\arabic*)., leftmargin=*, itemsep=0.2em, topsep=0.2em]
    \item AIA \cite{Lemen_SoPh_2012} level 1 images in nine wavelengths:
    \begin{itemize}[label=--, leftmargin=*, itemsep=0.1em, topsep=0.1em]
        \item Extreme ultraviolet (EUV): 94 \AA, 131 \AA, 171 \AA, 193 \AA, 211 \AA, 304 \AA, and 335 \AA
        \item Ultraviolet (UV): 1600 \AA
        \item Visible light: 4500 \AA
    \end{itemize}
    \item High-resolution (1K) magnetograms from the HMI \cite{Schou_SoPy_2012}, also obtained from JSOC.
\end{enumerate}
We used the long-wavelength channel (1–8 Å) X-ray flux data from the GOES X-ray Sensor for class labels. 
Specifically, we collected Science-level data from GOES-15 for the period 2011-2020 and from GOES-16 for 2021-2022.

\section{Experimental Setup}
\label{supp:setup}
\subsection{Implementation Details}
\paragraph{Deep SWM.}
Table \ref{tab:experimental_setup} illustrates the experimental settings of the proposed method.
Our model had approximately 1.59M trainable parameters and 4.64G multiply-add operations. 
The proposed model was trained on an Nvidia H200 with 140GB of GPU memory and an Intel Xeon PLATINUM 8580 processor.
It took approximately three hours to train our model. The inference time was approximately 12 ms per sample.

We used the training set to train our model, the validation set for tuning the hyperparameters, and the test set for evaluating the model's performance. 
We computed the GMGS score on the validation set every epoch. The performance on the test set was given by the model that achieved the highest GMGS score on the validation set.

\paragraph{Sparse MAE.}
Sparse MAE, as described in Section \ref{sec:met-smae}, uses an encoder-decoder architecture with a reconstruction loss, similar to the original MAE \cite{He_MAE_2022}. Here, we provide details of the specific configurations used in our implementation.

The encoder of our pretraining model, composed of $L_\mathrm{enc}$ Transformer layers, is trained to encode the masked input resulting from the two-phase masking process applied to $\mathbf{V}_t$ into an intermediate feature representation. This representation is then processed by a decoder consisting of $L_\mathrm{dec}$ Transformer layers, which aims to reconstruct the original image. The reconstruction loss is calculated as the mean squared error between the original image and the reconstructed image. Following \cite{He_MAE_2022}, the loss is computed only over the masked pixels.

The experimental setup for MAE pretraining is shown in Table \ref{tab:experimental_setup_pretraining}. For MAE pretraining, the number of trainable parameters and the number of multiply-accumulate operations for the proposed method are 2.56M and 27.68G, respectively. For MAE pretraining, we followed the same procedure using the training, and validation sets.

\subsection{Preprocessing and Data Augmentation}
\label{supp:sub-preprocess}
We pre-processed the datasets by performing the following steps sequentially:
\begin{enumerate}[label=\arabic*., leftmargin=*, itemsep=0.2em, topsep=0.2em]
    \item Resize all images from $1024 \times 1024$ to $256 \times 256$ to reduce computational complexity.
    \item For HMI images, mask the timestamp information in the bottom-left corner.
    \item To align the solar scales between AIA and HMI images, crop the edges of the AIA images, followed by resizing to match the HMI resolution using bilinear interpolation.
    \item Perform standardization on each channel of all images in the dataset.
    \item Synchronize the HMI images, AIA images, and class labels to a 1-hour cadence through temporal resampling to maintain a consistent time series.
\end{enumerate}
For data augmentation, we applied random rotations, scaling, brightness and contrast adjustments, Gaussian blur, and channel-specific noise addition.

\subsection{Classifier Re-training}
Given the class imbalance in our dataset, sampling is necessary during training to ensure an equal number of samples across classes. 
However, oversampling can lead to overfitting, especially for the X-class, which has few samples.
Consequently, our proposed method incorporates a two-stage approach. 
In the first stage, we train the model on the original dataset.
In the second stage, we perform model retraining based on Classifier Re-training\cite{Kang_CRT_2020}. 
This two-stage process mitigates overfitting while addressing the class imbalance.

\subsection{Evaluation Metrics}
\label{supp:sub-metrics}
As described in Section \ref{sec:exp-eva}, we use GMGS, BSS, and TSS to evaluate the performance of our model. These metrics are defined below. 

The GMGS is defined as the trace of the product of a scoring matrix \(S\) and a contingency table \(P\):
\begin{equation}
\text{GMGS} = \text{tr}(S^T \cdot P),
\label{eq:gmgs}
\end{equation}
where \(S\) and \(P\) denote the \(I\)-rank scoring matrix with an element \(s_{ij}\) and the \(I\)-categorical contingency table with an element \(p_{ij}\), respectively. The GMGS is an important metric in recent solar flare prediction studies \cite{Gandin_gmgs_1992}. The elements \(s_{ij}\) of the symmetric scoring matrix \(S\) are defined as:
\begin{align}
s_{ii} &= \frac{1}{I-1} \left[ \sum_{k=1}^{i-1} a_k^{-1} + \sum_{k=i}^{I-1} a_k \right] \quad (1 \leq i \leq I),
\label{eq:sii} \\
\begin{split}
s_{ij} &= \frac{1}{I-1} \Biggl[ \sum_{k=1}^{i-1} a_k^{-1} + \sum_{k=i}^{j-1} (-1) + \sum_{k=j}^{I-1} a_k \Biggr] \\
&\qquad (1 \leq i < j \leq I),
\end{split}
\label{eq:sij}
\end{align}
\begin{align}
a_i &= \frac{1 - \sum_{k=1}^{i} p_k}{\sum_{k=1}^{i} p_k} \quad (1 \leq i \leq I),
\label{eq:ai} \\
p_i &= \sum_{j=1}^{I} p_{ij} \quad (1 \leq i \leq I).
\label{eq:pi}
\end{align}

\begin{figure}[tb!]
    \centering
    \includegraphics[width=\linewidth]{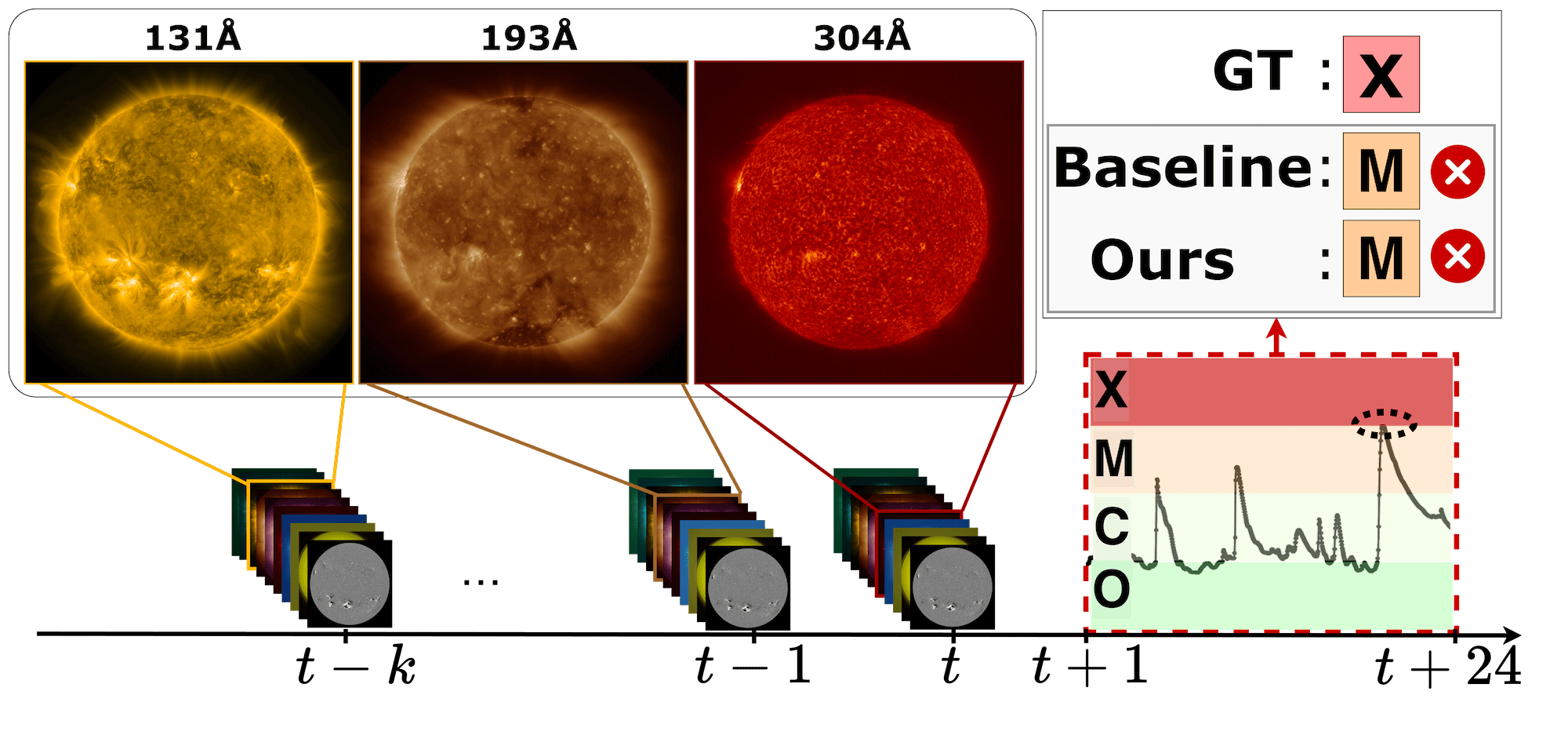}
    \caption{X-ray flux transitions leading up to an X1.0 flare event.}
    \label{fig:xray_transitions}
\end{figure}

\begin{figure*}[!htbp]
    \centering
    \includegraphics[width=\linewidth]{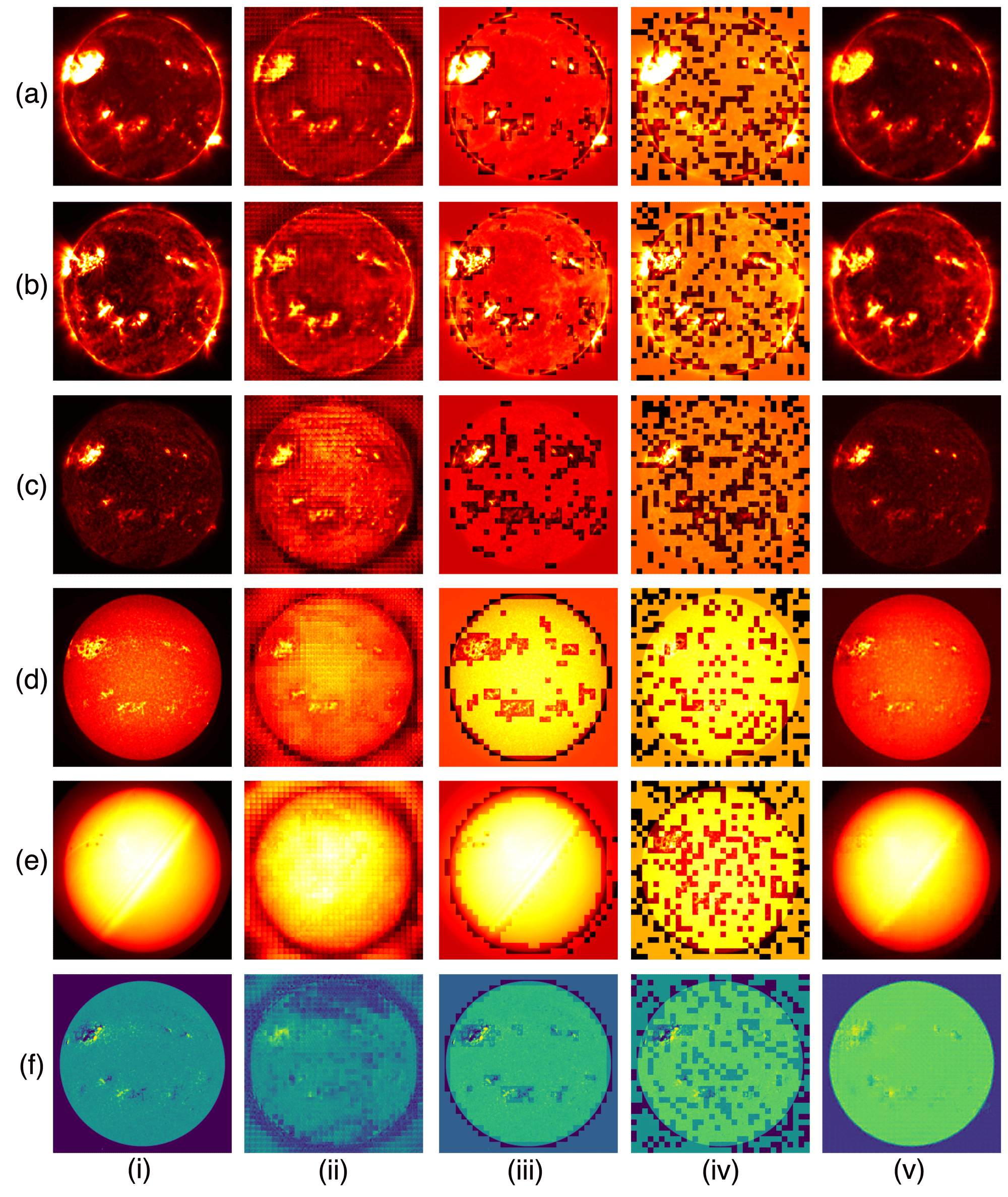}
    \caption{Reconstruction results obtained from the baseline MAE \cite{He_MAE_2022} ($\rho=0.5$) and our proposed Sparse MAE. Rows (a), (b), (c), (d), (e), and (f) show 94 Å, 171 Å, 304 Å, 1600 Å, 4500 Å AIA, and HMI images, respectively, captured three hours before an upcoming X-class flare. Columns (i), (ii), (iii), (iv), and (v) present the original image, the baseline reconstruction, a visualization of patches with the top $\alpha\%$ highest standard deviation highlighted, the spatial-level masking of the Sparse MAE, and the reconstruction of the Sparse MAE, respectively.}
    \label{fig:qualitative_results_pretraining_supp}
\end{figure*}

\input{tab/ablation_pretraining_appendix}
\input{tab/ablation_ssm}

The BSS, a standard metric for evaluating the reliability of solar flare forecasts \cite{Nishizuka_bss_2020}, is defined as:
\begin{align}
\text{BSS} &= \frac{\text{BS} - \text{BS}_c}{0 - \text{BS}_c},
\label{eq:bss} \\
\text{BS} &= \sum_{n=1}^{N} \sum_{i=1}^{I} (p(\hat{y}_{ni}) - y_{ni})^2,
\label{eq:bs} \\
\text{BS}_c &= \sum_{n=1}^{N} \sum_{i=1}^{I} (f - y_{ni})^2,
\label{eq:bsc}
\end{align}
where \(N\), \(I\), \(y_{ni}\), \(p(\hat{y}_{ni})\), and \(f\) are the number of samples, the number of classes, the true label of the \(i\)-th class for the \(n\)-th sample, the predicted probability of the \(i\)-th class for the \(n\)-th sample, and the climatological event rate, respectively.

The TSS is given by:
\begin{equation}
\text{TSS} = \frac{\text{TP}}{\text{TP} + \text{FN}} - \frac{\text{FP}}{\text{FP} + \text{TN}},
\label{eq:tss}
\end{equation}
where TP, FP, FN, and TN denote the number of true positives, false positives, false negatives, and true negatives in the contingency table, respectively.

\section{Additional Experiments}

\subsection{Performance of Human Forecasters}
\label{supp:human_forecasters}

The performance of human experts in daily solar flare forecasting operations from 2000 to 2015 was reported by Kubo et al. \cite{Kubo_humanperf_2017}, and is summarized in Table \ref{tab:quantitative_results}.
These human experts were engaged in the same forecasting problem as FlareBench: predicting the maximum solar flare class within a 24-hour period. To issue daily forecasts, they utilized solar indicators, including the current and historical X-ray flux, sunspot magnetic field configurations, and chromospheric brightenings in active regions, highlighting their expertise in operational solar flare prediction.

\subsection{X-ray Flux Transitions for a Challenging Case}
\label{supp:xray_transitions}
Fig. \ref{fig:qualitative_results} (d) illustrates a failed X-class prediction. Fig. \ref{fig:xray_transitions} shows the X-ray flux transitions for this event, including the 24-hour period leading up to the X1.0 flare. Appendix \ref{supp:flare_class} defines the flare classes, including the notation where a number follows the class letter (e.g., X1.0, M2.5). During the 24 hours preceding the X-class flare, there were two distinct peaks in X-ray flux, each corresponding to M-class flares. Furthermore, the X-class flare represented the boundary between X-class and M-class flares. These factors likely contributed to the difficulty in accurately classifying the flare class. 

\subsection{Qualitative Results for Pretraining}
\label{supp:qualitative_pretraining}
Fig.~\ref{fig:qualitative_results_pretraining_supp} illustrates the reconstruction results obtained from the baseline MAE \cite{He_MAE_2022} with a mask ratio ($\rho$) of 0.5 and our proposed Sparse MAE. Rows (a), (b), (c), (d), (e), and (f) show 94 Å, 171 Å, 304 Å, 1600 Å, 4500 Å AIA, and HMI images, respectively, captured three hours before an upcoming X-class flare. Columns (i), (ii), (iii), (iv), and (v) present the original image, the baseline reconstruction, a visualization of patches with the top $\alpha\%$ highest standard deviation highlighted, the spatial-level masking of the Sparse MAE, and the reconstruction of the Sparse MAE, respectively.

As depicted in subfigures (a-ii) and (a-v), and the others, the Sparse MAE reconstructs features in and around sunspots with high fidelity. By contrast,  the baseline method struggles to reproduce fine details in these regions.
These observations suggest that the enhanced representation of sunspots in the Sparse MAE reconstructions can be attributed to its two-phase masking strategy, which emphasizes preserving essential features such as sunspots.

\subsection{Additional Ablation Study}
\label{supp:ablation_study}
\paragraph{Pretraining ablation.}
Table \ref{tab:ablation_pretraining_appendix} presents an ablation study comparing the impact of different pretraining methods on solar flare prediction performance. We evaluate three models: (2-i) using MAE with a mask ratio ($\rho$) of 0.75, (2-ii) using MAE with $\rho$ = 0.5, and (2-iii) our proposed Sparse MAE.
The table presents the GMGS, BSS$_{\scalebox{0.8}{$\geq \mathrm{M}$}}$, and TSS$_{\scalebox{0.8}{$\geq \mathrm{M}$}}$ scores for the solar flare prediction task, and the MSE($r<0.65$) and MSE from the pretraining phase. MSE($r<0.65$) and MSE are computed as the mean squared error over the masked patches. MSE($r<0.65$) is restricted to patches within a defined solar region: a circle centered at the image center with a normalized radius of 0.65 (where the distance from the image center to a corner is normalized to 1.0). In this context, a normalized radius of 0.65 defines the boundary of the solar disk. We focus on MSE($r<0.65$) because accurate reconstruction of the solar disk is more critical for solar flare prediction than reconstruction of the surrounding non-solar regions.

The results reveal differences in performance across the evaluated models.  Models (2-i) and (2-ii) underperform Model (2-iii) in terms of both GMGS and MSE($r<0.65$). Specifically, Model (2-i) underperforms Model (2-iii) by 0.296 and 6.582 points in GMGS and MSE($r<0.65$), respectively, while Model (2-ii) underperforms Model (2-iii) by 0.162 and 4.328 points in GMGS and MSE($r<0.65$), respectively. These results indicate that the improved reconstruction of crucial solar regions, such as sunspots in sparse solar images, achieved with Sparse MAE, leads to extracting features more relevant for solar flare prediction.

\paragraph{Deep SSM ablation.}
Table \ref{tab:ablation_ssm} presents the performance impact of different architectures in the temporal modeling components.
We compared models using the following architectures for capturing temporal dependencies: (3-i) Attention \cite{Vaswani_NeurIPS_2017}, (3-ii) Mamba \cite{Gu_Mamba_2024}, and (3-iii) S5 \cite{Smith_S5_2023}. In our method, these architectures replace the ST-SSM, LT-SSM, and their integration mechanism.
From Table \ref{tab:ablation_ssm}, Models (3-i) and (3-ii) underperformed Model (3-iii) in GMGS by 0.271, 0.218 points, respectively.
These results suggest that S5 \cite{Smith_S5_2023} accurately captures temporal dependencies in solar images.

\subsection{Error Analysis}
\label{supp:error_analysis}
Table \ref{tab:confusion_matrix} presents the confusion matrix obtained using our method on the test set of the third fold.
Given the significant impact of X-class solar flares, our model prioritizes their accurate prediction. This prioritization results in more false positives for the X-class, as illustrated in the confusion matrix because correctly identifying these impactful events is paramount.

\input{tab/confusion_matrix}
\input{tab/error_analysis}
We defined samples that were incorrectly predicted as failure cases.
There were 8,832 failure cases identified in the third fold of the time-series cross-validation.
Table \ref{tab:error_analysis} categorizes the failed cases. We used the metric $\text{GMGS}_{\text{Influence}}$ (as introduced in \cite{Kaneda_FlareTransformer_2022}) to calculate the influence of failure cases on the GMGS. The influence for each element (i,j) of the confusion matrix is defined as:
\begin{equation}
\text{GMGS}_{\text{Influence}_{ij}} = \frac{c_{ij}(s_{ii} - s_{ij})}{N},
\label{eq:gmgs_influence}
\end{equation}
where $c_{ij}$, $s_{ij}$, and $N$ represent element $(i, j)$ of the confusion matrix, element $(i, j)$ of the GMGS score matrix (detailed in Subsection~\ref{sec:exp-eva}), and the total number of samples, respectively. This metric provides a quantitative measure of how much each type of error negatively impacts the overall GMGS.

Table \ref{tab:error_analysis} indicates that the bottleneck is the misclassification of C-class flares as X-class. Given the potential for severe consequences, the model prioritizes predicting X-class flares. This prioritization reduces the risk of missing true X-class events (false positives) but may increase false negatives, as shown in the confusion matrix (Table \ref{tab:confusion_matrix}). 
This trade-off is a deliberate design choice to reduce the likelihood of failing to identify high-impact events.

%% file: tab/flare_class.tex
\begin{table}[t]
\centering
\small
\vspace{-1mm}
\caption{Correspondence between flare classes and peak X-ray flux intensities}
\vspace{-1mm}
\label{tab:flare_class}
\begin{tabularx}{\linewidth}{>{\centering\arraybackslash}m{3cm} >{\centering\arraybackslash}X} 
\hline
\multicolumn{1}{c}{Flare Class} & \multicolumn{1}{c}{Peak X-ray Flux (I) [W/m\textsuperscript{2}]} \\
\hline
\hline
X & \(I > 10^{-4}\) \\
M & \(10^{-5} < I \leq 10^{-4}\) \\
C & \(10^{-6} < I \leq 10^{-5}\) \\
O & \(I \leq 10^{-6}\) \\
\hline
\end{tabularx}
\vspace{-2mm}
\end{table}

%% file: tab/experimental_setup.tex
\begin{table}[t]
  \centering
  \caption{Experimental setup for the proposed method.}
  \label{tab:experimental_setup}
  \begin{tabular}{ll}
    \toprule
    Epoch (first stage)          & 20 \\
    Epoch (second stage)         & 15 \\
    Batch size                   & 32 \\
    Optimizer                    & \begin{tabular}[c]{@{}l@{}}AdamW \\ 
                                  ($\beta_1=0.9,\;\beta_2=0.95$)\end{tabular} \\
    Learning Rate (first stage)  & $4.0 \times 10^{-5}$ \\
    Learning Rate (second stage) & $4.0 \times 10^{-5}$ \\
    Weight decay (first stage)   & $5.0 \times 10^{-2}$ \\
    Weight decay (second stage)  & $5.0 \times 10^{-2}$ \\
    $\#L_\mathrm{SSE}$                    & 3 \\
    $\#L_\mathrm{LT}$                     & 1 \\
    $\#D$                          & 64 \\
    $\#\lambda_\mathrm{CE}$                & 1 \\
    $\#\lambda_\mathrm{BSS}$               & 2 \\
    $\#\lambda_\mathrm{GMGS}$            & 1 \\
    $\#k$            & 4 \\
    $\#m$            & 672 \\
    \bottomrule
  \end{tabular}
\end{table}

%% file: tab/experimental_setup_pretraining.tex
\begin{table}[t]
  \centering
  \caption{Experimental setup for the Sparse MAE.}
  \label{tab:experimental_setup_pretraining}
  \begin{tabular}{ll}
    \toprule
    Epoch              & 20 \\
    Batch size         & 32 \\
    Optimizer          & \begin{tabular}[c]{@{}l@{}}AdamW \\ 
                         ($\beta_1=0.9,\;\beta_2=0.95$)\end{tabular} \\
    Learning Rate      & $4.0 \times 10^{-3}$ \\
    Weight decay       & $5.0 \times 10^{-2}$ \\
    $\#\alpha$ & 20 \\
    $\#L_{\mathrm{enc}}$ & 8 \\
    $\#L_{\mathrm{dec}}$ & 12 \\
    $\#D_{\mathrm{pre}}$ & 128 \\
    $\#r_l$ & 0.3\\
    $\#r_h$ & 0.5 \\
    $\#r_f$ & 0.5 \\
    \bottomrule
  \end{tabular}
\end{table}

%% file: tab/ablation_pretraining_appendix.tex
\begin{table*}[htbp]
  \centering
  \normalsize
  \setlength{\tabcolsep}{4pt}
  \renewcommand{\arraystretch}{1.2}
  \scalebox{0.9}{%
    \begin{tabular}{c l c c c c c}
      \toprule
      Model & Masking & GMGS$\uparrow$ & BSS$_{\geq\text{M}}$$\uparrow$ & TSS$_{\geq\text{M}}$$\uparrow$ & MSE ($r<0.65$)$\downarrow$ & MSE$\downarrow$ \\
      \hhline{=======}
      (1-i) & \makecell{MAE \cite{He_MAE_2022} \linebreak ($\rho$=0.75)} &
      $0.286 \pm \scriptstyle 0.090$ &
      $0.067 \pm \scriptstyle 0.306$ &
      $0.428 \pm \scriptstyle 0.173$ &
      $9.837 \pm \scriptstyle 0.365$ &
      $4.147 \pm \scriptstyle 0.378$ \\
      (1-ii) & \makecell{MAE \cite{He_MAE_2022} \linebreak ($\rho$=0.5)} &
      $0.420 \pm \scriptstyle 0.062$ &
      $\mathbf{0.354 \pm \scriptstyle 0.163}$ &
      $0.402 \pm \scriptstyle 0.100$ &
      $7.583 \pm \scriptstyle 0.372$ &
      $5.887 \pm \scriptstyle 0.329$ \\
      \rowcolor{red!10}
      (1-iii) & Sparse MAE (Ours) &
      $\mathbf{0.582 \pm \scriptstyle 0.032}$ &
      $0.334 \pm \scriptstyle 0.299$ &
      $\mathbf{0.543 \pm \scriptstyle 0.074}$ &
      $\mathbf{3.255 \pm \scriptstyle 0.180}$ &
      $\mathbf{2.461 \pm \scriptstyle 0.115}$ \\
      \bottomrule
    \end{tabular}
  }
    \vspace{-3mm}
  \caption{Ablation study: impact of masking strategies.}
  \vspace{-5mm}
  \label{tab:ablation_pretraining_appendix}
  
\end{table*}

%% file: tab/ablation_ssm.tex
\begin{table}[t]
\centering
\footnotesize 
\setlength{\tabcolsep}{3pt} 
\renewcommand{\arraystretch}{1.2} 
\resizebox{\columnwidth}{!}{
\begin{tabular}{ccccc}
\toprule
Model & Architecture & GMGS$\uparrow$ & BSS$_{\geq\text{M}}$$\uparrow$ & TSS$_{\geq\text{M}}$$\uparrow$ \\
\hhline{=====} 
(3-i) & Attention \cite{Vaswani_NeurIPS_2017}
    & 0.311 $\pm$ 0.177
    & \textbf{0.753 $\pm$ 0.117}
    & 0.287 $\pm$ 0.151 \\
(3-ii) & Mamba \cite{Gu_Mamba_2024}
     & 0.364 $\pm$ 0.070
     & 0.663 $\pm$ 0.032
     & 0.364 $\pm$ 0.087 \\
\rowcolor{red!10}
(3-iii) & S5 \cite{Smith_S5_2023}
      & \textbf{0.582 $\pm$ 0.032}
      & 0.334 $\pm$ 0.299
      & \textbf{0.543 $\pm$ 0.074} \\
\bottomrule
\end{tabular}
}
\vspace{-3mm}
\caption{Performance comparison of different architectures.}
\label{tab:ablation_ssm}
\vspace{-5mm}
\end{table}

%% file: tab/confusion_matrix.tex
\begin{table}[]
\centering
\begin{tabular}{|cl|llll|}
\hline
\multicolumn{1}{|l}{} &
  & \multicolumn{4}{l|}{Predicted Flare Class} \\
\cline{3-6}
\multicolumn{1}{|l}{} &
  & O & C & M & X \\
\hline
\multicolumn{1}{|c|}{\multirow{4}{*}{\makebox[2.5cm][c]{Observed flare class}}}
  & O & 5953 & 1110 & 29   & 167  \\
\multicolumn{1}{|c|}{} 
  & C & 1427 & 2307 & 1211 & 2329 \\
\multicolumn{1}{|c|}{} 
  & M & 111  & 321  & 433  & 1394 \\
\multicolumn{1}{|c|}{} 
  & X & 10   & 32   & 30   & 139  \\
\hline
\end{tabular}
\vspace{-3mm}
\caption{Confusion matrix for the test set of the third fold.}
\label{tab:confusion_matrix}
\end{table}

%% file: tab/error_analysis.tex
\begin{table}[t]
  \centering
  \begin{tabular}{lll}
    \hline
    Observed class & Predicted class & GMGS\_influence \\ \hline\hline
    \rowcolor{gray!10} C & X & 0.1007 \\
    C & O & 0.0560 \\
    \rowcolor{gray!10} O & C & 0.0435 \\
    C & M & 0.0281 \\
    \rowcolor{gray!10} M & X & 0.0278 \\ \hline
  \end{tabular}
\vspace{-2mm}
  \caption{The error analysis on the third fold's test set using the $\text{GMGS}_{\text{Influence}}$.}
  \vspace{-4mm}
  \label{tab:error_analysis}
\end{table}